\PassOptionsToPackage{sort}{natbib}

\documentclass[preprint, 3p, times, twocolumn]{elsarticle}

\usepackage{amssymb}
\usepackage{amsmath}
\usepackage{acronym}
\usepackage{algorithm}
\usepackage{algorithmic}
\usepackage{makecell}
\usepackage[inline]{enumitem}
\usepackage{bbm}
\usepackage[skip=1pt]{caption}
\usepackage{natbib}
\usepackage{xcolor}
\usepackage{booktabs}
\usepackage{hyperref}
\usepackage{rotating}
\usepackage{float}
\usepackage{multirow}
\usepackage{subfigure}

\newcommand{\rev}[1]{{#1}}
\newcommand{\revv}[1]{{#1}}

\acrodef{RMSE}{root mean squared error}

\makeatletter
\def\ps@pprintTitle{%
  \let\@oddhead\@empty
  \let\@evenhead\@empty
  \def\@oddfoot{\reset@font\hfil\thepage\hfil}
  \let\@evenfoot\@oddfoot
}
\makeatother

\begin{document}

\begin{frontmatter}

\title{Hierarchical Forecasting at Scale}

\author[1]{Olivier~Sprangers}
\ead{o.r.sprangers@uva.nl}

\author[2]{Wander~Wadman}
\ead{wwadman@bol.com}

\author[3]{Sebastian~Schelter}
\ead{s.schelter@uva.nl}

\author[3]{Maarten~de~Rijke}
\ead{m.derijke@uva.nl}

\address[1]{AIRLab, University of Amsterdam, Science Park 900, 1098 XH Amsterdam, The Netherlands}
\address[2]{bol., Papendorpseweg 100, 3528 BJ Utrecht, The Netherlands}
\address[3]{University of Amsterdam, Science Park 900, 1098 XH Amsterdam, The Netherlands}

\begin{abstract}
  Hierarchical forecasting techniques allow for the creation of forecasts that are coherent with respect to a pre-specified hierarchy of the underlying time series. This targets a key problem in e-commerce, where we often find millions of products across many product hierarchies, and forecasts need to be made for both individual products and product aggregations. However, existing hierarchical forecasting techniques scale poorly when the number of time series increases, which limits their applicability at a scale of millions of products. 
  
  In this paper, we propose to learn a coherent forecast for millions of products with a single bottom-level forecast model by using a loss function that directly optimizes the hierarchical product structure. We implement our loss function using sparse linear algebra, such that the number of operations in our loss function scales quadratically rather than cubically with the number of products and levels in the hierarchical structure. The benefit of our sparse hierarchical loss function is that it provides practitioners a method of producing bottom-level forecasts that are coherent to any chosen cross-sectional or temporal hierarchy. In addition, removing the need for a post-processing step as required in traditional hierarchical forecasting techniques reduces the computational cost of the prediction phase in the forecasting pipeline, as well as its deployment complexity.  
  
  In our tests on the public M5 dataset, our sparse hierarchical loss function performs up to 10\% better as measured by RMSE and MAE than the baseline loss function. Next, we implement our sparse hierarchical loss function within an existing gradient boosting-based forecasting model at \textit{bol}, a large European e-commerce platform. At \textit{bol}, each day a forecast for the weekly demand of every product for the next twelve weeks is required. In this setting our sparse hierarchical loss resulted in an improved forecasting performance as measured by RMSE of about 2\% at the product level, as compared to the baseline model, and an improvement of about 10\% at the product level as measured by MAE. Finally, we found an increase in forecasting performance of about 5--10\% (both RMSE and MAE) when evaluating the forecasting performance across the cross-sectional hierarchies that we defined. These results demonstrate the usefulness of our sparse hierarchical loss applied to a production forecasting system at a major e-commerce platform.

\end{abstract}


\end{frontmatter}

\section{Introduction} \label{sec:intro}
In e-commerce, we are often faced with two forecasting challenges. First, forecasts at the lowest granularity -- often the individual product level -- are required but we also need forecasts at higher granularities, for example at the category, department, or regional level, as higher level forecasts are often needed in logistics and financial planning. Second, forecasts at different time granularities are required, for example daily or weekly forecasts. It is common that separate forecast models are made for each separate (temporal) granularity, and as such these forecasts may not be coherent with each other. Hierarchical forecasting \cite{hyndman_optimal_2011} and temporal hierarchical forecasting techniques \cite{athanasopoulos_forecasting_2017,rangapuram_coherent_2023,theodosiou_forecasting_2021} aim to solve the problem of creating forecasts that are coherent with respect to a pre-specified cross-sectional and/or temporal hierarchy of the underlying time series. 

\paragraph{Challenges with existing cross-sectional and temporal hierarchical forecasting techniques} Reconciliation methods adjust the forecasts for each level in the hierarchy by minimizing the errors at each forecast level. These methods are applied as a post-processing step that requires a matrix inversion that scales cubically with the number of products or product hierarchies \cite{hyndman_optimal_2011,athanasopoulos_forecasting_2017,wickramasuriya_optimal_2019}. In settings with millions of products such as in e-commerce, this becomes computationally expensive at prediction time. Neural network methods can optimize for the hierarchy in an end-to-end manner, however, these are either multivariate methods that scale poorly to millions of time series \cite{rangapuram_endtoend_2021} or they can only optimize for the temporal hierarchy \cite{rangapuram_coherent_2023}.
  
\paragraph{Sparse loss function} In order to overcome these scaling issues, we design a sparse \emph{hierarchical loss} (HL) function that directly optimizes both cross-sectional and temporal hierarchical structures. Our corresponding sparsity-aware implementation ensures that the number of operations in our loss function scales quadratically rather than cubically with the number of products and levels in the hierarchical structure, enabling computationally efficient training. The benefit of our sparse hierarchical loss function is that it provides practitioners a method of producing bottom-level forecasts that are coherent to any chosen cross-sectional and temporal hierarchy. In addition, removing the need for a post-processing step as used in traditional hierarchical forecasting techniques reduces the computational cost of the prediction phase in the forecasting pipeline. Furthermore, this also reduces the deployment complexity of the forecasting pipeline.

\paragraph{Evaluation} We evaluate our sparse HL function on a gradient-boosted forecasting system on the public M5 dataset \cite{makridakis_m5_2022} and a proprietary dataset from our e-commerce partner. For the M5 dataset, we demonstrate that our implementation provides up to 10\% better forecasting performance as measured by both RMSE and MAE compared with (i) reconciliation methods and (ii) baseline bottom-level forecasting methods that use a standard loss function. For the proprietary dataset, we present the results of an offline test on the product-level forecast system of \textit{bol}, a European e-commerce company with a catalog of millions of unique products. We find our sparse HL function improves the forecasting performance by about 2\% on RMSE and 10\% on MAE as compared to the baseline forecasting system. This demonstrates the usefulness of our sparse HL function in a large-scale setting.

\paragraph{Contributions} In summary, the main contributions of this paper are:
\begin{enumerate}
  \item We design a sparse hierarchical loss function that enables direct end-to-end training of cross-sectional and temporal hierarchical forecasts in large-scale settings in Section~\ref{sec:sparsehloss}.
  \item We empirically demonstrate that our sparse hierarchical loss function can outperform existing hierarchical forecasting reconciliation methods by up to 10\% in Section~\ref{subsec:publicdatasets}. \rev{Contrary to most end-to-end hierarchical forecasting methods that leverage neural networks \cite{rangapuram_endtoend_2021,rangapuram_coherent_2023}, we use LightGBM \cite{ke_lightgbm_2017} as our base forecasting model, a highly popular gradient boosting-based forecasting method that is widely used in industry \cite{januschowski_forecasting_2022} and was used by the majority of the top performing solutions in the M5 forecasting competition \cite{makridakis_m5_2022}.}
  \item We show how our sparse hierarchical loss function scales to large-scale settings and demonstrate a reduction of both training and prediction time of up to an order of magnitude compared to the best hierarchical forecasting reconciliation methods (Section~\ref{par:timecomplexity}).
  \item We present the results of an offline test of our method for the primary product demand forecasting model at \textit{bol}, a European e-commerce company with a catalogue of millions of unique products, demonstrating an improvement of 2\% on RMSE and 10\% on MAE as compared to the baseline forecasting system, in Section~\ref{subsec:proprietarydatasets}.
\end{enumerate}

\section{Related work} \label{sec:relwork}

\paragraph{Forecasting for large-scale settings} Contemporary large-scale forecasting applications require forecasting many time series concurrently \cite{bose_probabilistic_2017}. In academia, there has been a surge in the use of neural network-based forecasting methods, which are methods that commonly learn a single forecast model that can produce forecasts for many time series. We refer the interested reader to the recent survey of \citet{benidis_deep_2023} for an overview of these methods. However, tree-based methods topped the M5 forecasting competition \cite{makridakis_m5_2022}, which is believed to be due to the strong implementations available of these algorithms \cite{januschowski_forecasting_2022}, such as the LightGBM \cite{ke_lightgbm_2017} or XGBoost \cite{chen_xgboost_2016} packages. Our own experience within \textit{bol} confirms this view: the ease of use, execution speed and strong default performance are key reasons a tree-based method is often the default choice when creating a new forecasting model.

\paragraph{Hierarchical forecasting} Hierarchical forecasting \cite{hyndman_optimal_2011, hyndman_fast_2016, ben_taieb_coherent_2017, bentaieb_regularized_2019, wickramasuriya_optimal_2019} and temporal hierarchical forecasting techniques \cite{ben_taieb_sparse_2017,athanasopoulos_forecasting_2017,rangapuram_coherent_2023,theodosiou_forecasting_2021} aim to solve the problem of creating forecasts that are coherent with respect to a pre-specified cross-sectional and/or temporal hierarchy of the underlying time series. We divide hierarchical forecasting methods into \textit{Reconciliation methods} and \textit{Other methods}. 

\textit{Reconciliation methods}. \rev{For a detailed overview of reconciliation methods, we refer the interested reader to the recent survey of \citet{athanasopoulos_forecast_2024}.} Reconciliation methods solve the hierarchical forecasting problem as a post-processing step by reconciling the forecasts to a pre-specified cross-sectional and/or temporal hierarchy \cite{hyndman_optimal_2011, hyndman_fast_2016, ben_taieb_coherent_2017, bentaieb_regularized_2019, wickramasuriya_optimal_2019, panagiotelis_forecast_2021, girolimetto_point_2023}. Limitations of these approaches are (i) they require a post-processing step, (ii) computing the reconciliation may be computationally expensive, as we show in Section~\ref{subsec:ourwork}, and (iii) approaches that are computationally less expensive tend to perform worse, as we show in Section~\ref{sec:experiments}. Recent work by \citet{ben_taieb_sparse_2017} and \citet{bentaieb_regularized_2019} has improved forecasting performance of previous reconciliation approaches but at the expense of even higher computational costs, as we explain in Section~\ref{sec:background}.

\textit{Other methods}. In \cite{rangapuram_endtoend_2021,rangapuram_coherent_2023} neural network-based end-to-end hierarchical probabilistic forecasting method are proposed to solve the hierarchical forecasting problem. More recently and most closely related to our work, \citet{han_simultaneously_2021} introduced SHARQ, a method that reconciles probabilistic hierarchical forecasts during training by employing a regularized loss function that aims to improve hierarchical consistency of bottom-up forecasts through regularization. However, the regularization does not strictly enforce the cross-sectional hierarchy in this method.

\section{Background} \label{sec:background}
To understand our problem setting and the issues we identify with existing hierarchical forecasting methods, we introduce the hierarchical forecasting problem and common methods of solving the hierarchical forecasting problem. 

\subsection{Problem definition}
\rev{Suppose we have \(n\) time series written as \(\textbf{y}_t \in \mathbb{R}^{n}\), where \(t\) denotes the time stamp. We are interested in finding \(h\)-step ahead estimates \(\hat{\textbf{y}}_{h}\) of the time series \(\textbf{y}_{T + h}\) using past values \(\textbf{y}_{1}, \dots, \textbf{y}_{T}\).}
In our hierarchical forecasting setting, we aim to create forecasts for many time series concurrently, whilst adhering to pre-specified hierarchical relationships that exist between the time series. This can be formalized as follows \rev{\cite{hyndman_forecasting_2021, athanasopoulos_forecast_2024}}:
\rev{\begin{equation} \label{eq:hfp}
  \tilde{\textbf{y}}_{h} = S G \hat{\textbf{y}}_{h} \;,
\end{equation}}
where \rev{\(S \in \{0, 1\}^{n \; \times \; n_b}\) is a matrix that defines the hierarchical relationship between the \(n_b\) bottom-level time series and the \(n_a = n - n_b\) aggregations, \(G \in \mathbb{R}^{n_b \; \times \; n}\)} is a matrix that encapsulates the contribution of each forecast to the final estimate, and \rev{\(\tilde{\textbf{y}}_{h} \in \mathbb{R}^{n} \)} is the vector of forecasts adjusted for the hierarchy. We can use the matrix \rev{\(G\)} to define various forecast contribution scenarios. Note that we can straightforwardly extend Eq.~\eqref{eq:hfp} to the setting of \textit{temporal hierarchies} \cite{athanasopoulos_forecasting_2017,rangapuram_coherent_2023} by considering forecasts of different time granularities in our vector of base forecasts \rev{\(\hat{\textbf{y}}_{h}\)} and using an appropriate choice of \(S\) to aggregate series of a different time granularity. We will show how cross-sectional and temporal hierarchical forecasting can be combined in Section~\ref{sec:sparsehloss}.

The optimal solution to the problem in Eq.~\eqref{eq:hfp} can be found using \textit{Reconciliation methods} and \textit{Other methods}.

\paragraph{Reconciliation methods}

\rev{\textit{MinTShrink} \cite{wickramasuriya_optimal_2019, athanasopoulos_forecast_2024} and variants find the optimal \rev{\(G\)} matrix by solving a minimization problem that has the following solution (ref. Theorem 1 of \cite{wickramasuriya_optimal_2019})}:
\rev{\begin{align} 
  G &= (J - JWU(U^TWU)^{-1}U^T) \;, \label{eq:p1}
\end{align}}
in which \(S\) is partitioned as \rev{\(S^T = [C^T \; I_{n_b}]\), \(J = [0_{n_b \; \times \; n_a} \; I_{n_b}]\), \(U^T = [I_{n_a} \; -C]\)}. In \textit{MinTShrink}, \(W\) is estimated using the shrunk empirical covariance estimate of \cite{schafer_shrinkage_2005}. Simpler choices for \(W\), such as the identity matrix, reduce the solution to the \textit{Ordinary Least Squares (OLS)} solution of \cite{hyndman_optimal_2011}. In \textit{ERM}, \citet{bentaieb_regularized_2019} note than \textit{MinTShrink} and variants rely on the assumption of unbiasedness of the base forecasts. They relax this assumption by formulating the hierarchical reconciliation problem as an \textit{Empirical Risk Minimization} problem, introducing the \textit{ERM} method. In addition, they propose two regularized variants of \textit{ERM} aimed at reducing forecast variance. 

\paragraph{Other methods} \textit{Hier-E2E} \cite{rangapuram_endtoend_2021} solves the problem of Eq.~\eqref{eq:hfp} by learning a neural network model that combines the forecasting and reconciliation step in a single model, resulting in an end-to-end solution removing the need for a post-processing step. Similarly, \textit{COPDeepVAR} \cite{rangapuram_coherent_2023} is an end-to-end neural network method that enforces temporal hierarchies, however this is a univariate method that is not able to enforce structural hierarchies (i.e., cross-sectional hierarchies) simultaneously, and therefore not suited to our task. \textit{SHARQ} \cite{han_simultaneously_2021} also moves the reconciliation step into the training phase and achieves reconciliation using a regularized loss function, where the regularization enforces the coherency. However, this method does not enforce absolute coherency to the hierarchy.

\subsection{Scaling issues of hierarchical forecasting methods} \label{subsec:ourwork}

Our main motivation for this paper are the limitations of prior work for problem settings with many time series.

\paragraph{Scaling issues with reconciliation methods} \label{sec:scalingissuesreconmethods}
In reconciliation methods, we encounter the following issues when scaling to many time series:
\begin{itemize}
  \item The reconciliation is performed as a \textit{post\--pro\-ces\-sing} step, and thus has to be performed as an additional step after generating the base forecasts. Even though \rev{\(G\)} in Eq.~\eqref{eq:hfp} needs to be computed only once using Eq.~\eqref{eq:p1}, the reconciliation still needs to be performed after each base forecast is produced. Also, \rev{\(G\)} ideally is sparse \cite{bentaieb_regularized_2019}, but no reconciliation method guarantees this, so computing Eq.~\eqref{eq:hfp} will generally be a dense matrix-vector product that scales with the number of time series.
  \item For \textit{MinTShrink} \cite{wickramasuriya_optimal_2019}, estimating \(W\) according to the method of \cite{schafer_shrinkage_2005} is computationally expensive, with a computational complexity of \rev{\(O(Nn^2)\)}, where \(N\) denotes the number of training samples used to compute the shrunk covariance estimate. In addition, the shrunk covariance estimate of \cite{schafer_shrinkage_2005} is not guaranteed to give consistent results in high-dimensional settings \cite{touloumis_nonparametric_2015}, making it less applicable for problem settings with many time series. Finally, the estimate for \(W\) will generally be a dense matrix, so we cannot make use of efficient sparse algorithms to solve Eq.~\eqref{eq:p1}. However, even for simpler, sparse choices of \(W\) (such as the identity matrix of \textit{OLS} \cite{hyndman_optimal_2011}), we still need to invert a matrix of size \rev{\(n_a \times n_a\)} in order to solve Eq.~\eqref{eq:p1}, which becomes computationally costly for problems with many aggregations, which naturally arise in retail forecasting scenarios. For example, for the M5 retail forecasting competition \cite{makridakis_m5_2021}, \rev{\(n_a=12,350\)}, even though there are only 3,049 unique products in this dataset.  
  \item For \textit{ERM} and its regularized variants \cite{bentaieb_regularized_2019}, we need to either invert multiple dense matrices that scale quadratically with the number of time series, or we need to compute a Kronecker product that scales quadratically with the number of time series, followed by an expensive lasso search procedure. Improving the computational complexity of the \textit{ERM} methods is also mentioned in \cite{bentaieb_regularized_2019} as an avenue for future work.
\end{itemize}

\paragraph{Scaling issues with other methods} \label{sec:scalingissuesneuralmethods} \textit{Hier-E2E} \cite{rangapuram_endtoend_2021} is a multivariate method, which means both input and output of the neural network scale with the number of time series. For neural networks, this significantly adds to the training cost and parameter cost as a large number of parameters are required to handle all the separate time series. This in turn requires GPUs with more memory to train these models, which increases cost to operate them. 

\section{Sparse Hierarchical Loss} \label{sec:sparsehloss}
In this section we present our main technical contribution, the sparse hierarchical loss. First, we show how cross-sectional and temporal hierarchical forecasting can be combined. Then, we introduce our loss function and demonstrate it via a toy example.

\paragraph{Combining cross-sectional and temporal hierarchical forecasting} We are interested in finding forecasts that can be aggregated according to a pre-specified cross-sectional hierarchy \rev{\(S^{cs} \in \{0, 1\}^{n^{cs} \; \times \; n_b^{cs}}\) and temporal hierarchy \(S^{te} \in \{0, 1\}^{n^{te} \; \times \; n_b^{te}}\):
\begin{align} 
  \tilde{\textbf{y}}^{cs}_{h} &= S^{cs} G^{cs} \hat{\textbf{y}}^{cs}_{h} \;, \label{eq:hfpbu_c_full} \\
  \tilde{\textbf{y}}^{te} &= S^{te} G^{te} \hat{\textbf{y}}^{te} \;. \label{eq:hfpbu_t_full}
\end{align}}
These equations can be interpreted as follows:
\begin{itemize}
  \item In Eq.~\eqref{eq:hfpbu_c_full}, we aggregate \rev{\(n^{cs}_b\)} bottom-level time series from the same \rev{forecast \(h\)} across a set of \rev{\(n^{cs} = n_b^{cs} + n_a^{cs}\)} cross-sectional aggregations.
  \item In Eq.~\eqref{eq:hfpbu_t_full}, we aggregate each time series consisting of \rev{\(n_b^{te}\)} timesteps across a set of \rev{\(n^{te} = n_b^{te} + n_a^{te}\)} temporal aggregations, hence we drop the subscript \rev{\(h\)}.
\end{itemize}
We will only create bottom-level forecasts, thus \rev{\(G^{cs} = [0_{n_b^{cs}} \times n_a^{cs}  \; I_{n_b^{cs}}] \)} and \rev{\(G^{te} = [0_{n_b^{te}} \times n_a^{te} \; I_{n_b^{te}}] \)}, yielding:
\rev{
\begin{align} 
  \tilde{\textbf{y}}^{cs}_{h} &= S^{cs} \hat{\textbf{y}}^{n_b^{cs}}_{h} \;, \label{eq:hfpbu_c} \\
  \tilde{\textbf{y}}^{te} &= S^{te} \hat{\textbf{y}}^{n_b^{te}} \;, \label{eq:hfpbu_t}
\end{align}
where \(\hat{\textbf{y}}{}^{n_b^{cs}}_{h}\) and \(\hat{\textbf{y}}{}^{n_b^{te}}\) denote the bottom-level base forecasts for the cross-sectional and temporal hierarchies, respectively.}
Considering only bottom-level forecasts has a number of benefits: (i) each forecast is coherent to any hierarchy by design, and (ii) we reduce the number of required forecasts from \rev{\(n\)} to \rev{\(n_b\)}, which can be a significant reduction (there is no need for a forecast for \rev{\(n_a\)} aggregations in the hierarchy). 
We now construct a matrix of bottom-level base forecasts \rev{\(\hat{\textbf{Y}}{}^{n_b} \in \mathbb{R}^{n_b^{cs} \; \times \; n_b^{te}}\)}, in which the columns represent the forecasts of the bottom-level time series at a timestep \rev{\(h\)}. This allows us to combine \eqref{eq:hfpbu_c} and \eqref{eq:hfpbu_t} as follows:
\rev{\begin{equation}
  \tilde{\textbf{Y}} = S^{cs} \hat{\textbf{Y}}{}^{n_b} (S^{te})^\intercal \;, \label{eq:hfp_tc} 
\end{equation}}
in which \rev{\(\tilde{\textbf{Y}} \in \mathbb{R}^{n^{cs} \; \times \; n^{te}}\)} represents the matrix of forecasts aggregated according to both cross-sectional and temporal hierarchies. Equivalently, we can aggregate our bottom-level ground truth values \rev{\(\textbf{Y}^{n_b} \in \mathbb{R}^{n_b^{cs} \; \times \; n_b^{te}}\):
\begin{equation}
  \textbf{Y} = S^{cs} \textbf{Y}^{n_b} (S^{te})^\intercal \;. \label{eq:hfp_tc_gt} 
\end{equation}}

\paragraph{Sparse hierarchical loss} To find the best forecasts for the hierarchical forecasting problem \eqref{eq:hfp_tc}, we try to find a \rev{forecasting model using} gradient-based optimization of the following loss function:
\rev{\begin{align} 
  L &= \sum \left[ \frac{1}{2}\left(\left(\textbf{Y} - \tilde{\textbf{Y}}\right) \odot \left(\textbf{Y} - \tilde{\textbf{Y}}\right)\right) \oslash \left(d^{cs} d^{te} \right) \right]  \; , \label{eq:whloss}
\end{align}
 in which \(\sum\) denotes the sum over all \(n^{cs}\; \times \; n^{te}\) elements of the matrix contained in the summation, \(\odot\) denotes element-wise multiplication, \(\oslash\) denotes element-wise division, and the vectors \(d^{cs}\) and \(d^{te}\) read:
 \begin{align}
  d^{cs} &= l^{cs} S^{cs} \textbf{1}^{cs} \;, \\
  d^{te} &= \left(l^{te} S^{te} \textbf{1}^{te}\right)^\intercal \;,
 \end{align}
 where \(S^{cs} \textbf{1}^{cs}\) and \(S^{te} \textbf{1}^{te}\) denote the row-sum of \(S^{cs}\) and \(S^{te}\), respectively, and \(l^{cs}\) and \(l^{te}\) denote the number of levels in hierarchies \(S^{cs}\) and \(S^{te}\), respectively. We will detail the necessity of the element-wise division of Eq.~\eqref{eq:whloss} by the matrix \(\left(d^{cs} d^{te}\right)\) later in this section. Note that Eq.~\eqref{eq:whloss} shares similarities with the \textit{Weighted Root Mean Squared Error} from the M5 competition \cite{makridakis_m5_2022}.}

We can derive the gradient and \rev{the second-order derivative} of \eqref{eq:whloss} with respect to the bottom-level forecasts \rev{\(\hat{\textbf{Y}}{}^{n_b}\) (ref.~\ref{app:hl_derivation} for the full derivation): 
\begin{align} 
  \frac{\partial L}{\partial \tilde{\textbf{Y}}} &=  \left(\tilde{\textbf{Y}} - \textbf{Y}\right) \oslash \left( d^{cs} d^{te} \right) \;,  \label{eq:hfp_grad} \\
  \frac{\partial L}{\partial \hat{\textbf{Y}}{}^{n_b}} &= (S^{cs})^\intercal \left(\frac{\partial L}{\partial \tilde{\textbf{Y}}}\right) S^{te} \;, \label{eq:hfpbu_grad}  \\
  \frac{\partial^2 L}{\partial \left(\hat{\textbf{Y}}{}^{n_b} \right)^2} &= (S^{cs})^\intercal \left( \mathbf{1} \oslash \left(d^{cs} d^{te}\right) \right) S^{te} \;. \label{eq:hfpbu_hess}                                         
\end{align}}

\paragraph{Analysis} The best possible forecast is achieved when the loss \eqref{eq:whloss} is minimized, or equivalently when the gradient \eqref{eq:hfp_grad} is zero:
\rev{
\begin{align} 
  \frac{\partial L}{\partial \tilde{\textbf{Y}}} &= \left(\tilde{\textbf{Y}} - \textbf{Y}\right) \oslash \left(d^{cs} d^{te}\right) \;, \nonumber \\
                                                 &= \left( S^{cs} \hat{\textbf{Y}}{}^{n_b} (S^{te})^\intercal - S^{cs} \textbf{Y}^{n_b} (S^{te})^\intercal \right) \oslash \left(d^{cs} d^{te}\right) \;, \nonumber \\
                                                 &=  \left(S^{cs} \left(\hat{\textbf{Y}}{}^{n_b} - \textbf{Y}^{n_b} \right) (S^{te})^\intercal \right) \oslash \left(d^{cs} d^{te} \right) \;, \nonumber
\end{align}}
which becomes zero when \rev{\(\hat{\textbf{Y}}{}^{n_b} = \textbf{Y}{}^{n_b}\)}. Thus, the best forecast model is found when each bottom-level forecast equals the ground truth. This is equivalent to the standard (i.e., non-hierarchical) squared error loss often used in forecasting problems. We argue that our hierarchical loss gradient can be seen as a \textit{smoothed} gradient compared to the standard squared error loss gradient (i.e., \rev{\(\hat{\textbf{Y}}{}^{n_b} - \textbf{Y}{}^{n_b}\)}). For example, consider the canonical case where we have two bottom-level time series \rev{(\(n_b^{cs}=2\))} consisting of two timesteps \rev{(\(n_b^{te}=2\))}. Furthermore, suppose we have a single cross-sectional aggregation (the sum of the two time series, thus \rev{\(n_a^{cs} = 1\) and \(n^{cs} = n_a^{cs} + n_b^{cs} = 3\)}), and a single temporal aggregation (the sum of the two timesteps, thus \rev{\(n_a^{te} = 1\) and \(n^{te} = n_a^{te} + n_b^{te} = 3\)}). Finally, there are two levels in our cross-sectional hierarchy and in our temporal hierarchy, thus \rev{\(l^{cs} = 2\) and \(l^{te} = 2\)}. The standard squared error loss gradient for this problem is:
\begin{align} \label{eq:sqloss_grad}
  \begin{bmatrix}
    \frac{\partial L}{\partial \hat{\textbf{y}}_{0, 0}} & \frac{\partial L}{\partial \hat{\textbf{y}}_{0, 1}} \\
    \frac{\partial L}{\partial \hat{\textbf{y}}_{1, 0}} & \frac{\partial L}{\partial \hat{\textbf{y}}_{1, 1}}
    \end{bmatrix} 
    = 
\begin{bmatrix}
  e_{0, 0} & e_{0, 1} \\
  e_{1, 0} & e_{1, 1} \\
\end{bmatrix}  \;, 
\end{align}
in which \(e_{i,j}\) denotes the bottom-level \rev{forecast error (\(\hat{y}_{i,j} - {y}_{i,j}\)) of the \(i\)-th bottom-level timeseries and \(j\)-th timestep, respectively}.
For our hierarchical loss, Eq.~\eqref{eq:hfp_tc} reads:
\rev{
\begin{align}
  \tilde{\textbf{Y}} = \underbrace{
    \begin{bmatrix}
    1 &1 \\
    1 &0 \\
    0 &1
    \end{bmatrix}}_{S^{cs}}
    \underbrace{    
    \begin{bmatrix}
      \hat{\textbf{y}}_{0, 0} & \hat{\textbf{y}}_{0, 1} \\
      \hat{\textbf{y}}_{1, 0} & \hat{\textbf{y}}_{1, 1} \\
    \end{bmatrix}}_{\hat{\textbf{Y}}{}^{n_b}}
    \underbrace{
    \begin{bmatrix}
      1 &1 &0\\
      1 &0 &1\\
      \end{bmatrix}}_{(S^{te})^\intercal}     
    \;,
\end{align}}
and the gradient of the loss with respect to the bottom level time series Eq.~\eqref{eq:hfpbu_grad} reads (ref. \ref{app:hl_derivation_toy} for the full derivation):
\rev{
\begin{align*}
  \begin{split}
  \begin{bmatrix}
  \frac{\partial L}{\partial \hat{\textbf{y}}_{0, 0}} & \frac{\partial L}{\partial \hat{\textbf{y}}_{0, 1}} \\
  \frac{\partial L}{\partial \hat{\textbf{y}}_{1, 0}} & \frac{\partial L}{\partial \hat{\textbf{y}}_{1, 1}}
  \end{bmatrix} 
  = &\left[\begin{matrix}
    \frac{9}{16}e_{0, 0} + \frac{3}{16}e_{1, 0} + \frac{3}{16}e_{0, 1} + \frac{1}{16}e_{1, 1} \\
    \frac{9}{16}e_{1, 0} + \frac{3}{16}e_{0, 0} + \frac{3}{16}e_{1, 1} + \frac{1}{16}e_{0, 1} 
    \end{matrix} \right. \\
    &\quad \left.
    \begin{matrix}
      \frac{9}{16}e_{0, 1} + \frac{3}{16}e_{0, 0} + \frac{3}{16}e_{1, 1} + \frac{1}{16}e_{1, 0} \\
      \frac{9}{16}e_{1, 1} + \frac{3}{16}e_{1, 0} + \frac{3}{16}e_{0, 1} + \frac{1}{16}e_{0, 0} \\
      \end{matrix}\right] \; .
  \end{split}
  \nonumber         
\end{align*}}
When we compare this result to the standard squared error loss gradient Eq.~\eqref{eq:sqloss_grad}, we find that we \textit{smooth} the bottom-level gradient by adding to it portions of the gradients of all cross-sectional and temporal aggregations the bottom-level series belongs to. This derivation also shows the motivation of adding the denominator \rev{matrix \(\left( d^{cs} d^{te} \right)\)} to the loss function \eqref{eq:whloss}: it is neccessary to scale the aggregation gradients by the number of elements in the aggregation, otherwise the magnitude of the gradient grows with the number of time series and the number of levels in the hierarchy, which we found to be undesirable when trying to facilitate stable learning. \rev{Thus, we add (portions of) the average gradient of the aggregations to the bottom-level gradient.}
  
\paragraph{Sparsity} \(S^{cs}\) and \(S^{te}\) are highly sparse. For example, \(S^{cs}\) has \rev{at most \(n_b^{cs}l^{cs}\)} non-zero elements: the number of bottom-level time series multiplied by the number of aggregations in the hierarchy. Hence, the overall sparsity of \(S^{cs}\) is given by \( 1 - \frac{n_b^{cs}l^{cs}}{n^{cs}n_b^{cs}} \). For the M5 dataset \cite{makridakis_m5_2021}, \(n_b^{cs} = 3,049\), \(l^{cs} = 12\), \(n^{cs}=42,840\), corresponding to a sparsity of 99.97\%. Next, the matrix of bottom-level ground truth values \(\textbf{Y}{}^{n_b}\) in \eqref{eq:hfp_tc_gt} may be sparse too, for example in the case of products that are not on sale for every timestep \(n_b^{te}\) in the dataset. All these sources of sparsity motivate the use of sparse linear algebra when computing Eqs.~\eqref{eq:whloss}--\eqref{eq:hfpbu_hess}. 

\paragraph{Implementation} We implement the hierachical loss \eqref{eq:whloss}, the bottom-level gradient \eqref{eq:hfp_grad}, \eqref{eq:hfpbu_grad} and \rev{second-order derivative} \eqref{eq:hfpbu_hess} in Python using the sparse library from \textit{SciPy} \cite{virtanen_scipy_2020}. Note that Eqs.~\eqref{eq:hfp_grad}--\eqref{eq:hfpbu_grad} can be rearranged:
\rev{%
\begin{equation} 
  \frac{\partial L}{\partial \hat{\textbf{Y}}} = \frac{(S^{cs})^\intercal}{d^{cs}} \left( \tilde{\textbf{Y}} - \textbf{Y} \right) \frac{S^{te}}{d^{te}} \;,
\end{equation}}
such that the parts before and after the brackets can be precomputed as they do not depend on the forecast values \(\tilde{\textbf{Y}}\), avoiding a costly division operation inside a training iteration. Also note that the \rev{second-order derivative} \eqref{eq:hfpbu_hess} does not depend on the forecast values \(\tilde{\textbf{Y}}\), so it can be precomputed as well. Our implementation, including the code to reproduce the experiments on public data from Section~\ref{sec:experiments}, is available on GitHub.\footnote{\url{https://github.com/elephaint/hfas}}

\section{Experiments}
  \label{sec:experiments}
  In this section we empirically verify the usefulness of our sparse hierarchical loss. First, we evaluate forecasting accuracy using a set of experiments on the public M5 dataset \cite{makridakis_m5_2021}. Then, we evaluate our sparse hierarchical loss in an offline experiment on a proprietary dataset from our e-commerce partner.

  \subsection{Public datasets} \label{subsec:publicdatasets}
  \paragraph{Task \& dataset} Our task is to forecast product demand. We use the M5 dataset \cite{makridakis_m5_2021} for our offline, public dataset experiments. The M5 dataset contains product-level sales from Walmart for 3,049 products across 10 stores in the USA. Furthermore, the dataset contains 12 cross-sectional product aggregations (e.g. department, region), which allow us to test hierarchical forecasting performance. We preprocess the dataset resulting in a set of features as described in \ref{app:m5dataset}. We forecast 28 days into the future.
  
  \paragraph{Baseline models} For our baseline forecasting model, we primarily use LightGBM \cite{ke_lightgbm_2017}, trained to predict one-day ahead. We subsequently recursively generate predictions for 28 days. Tree-based models dominated the M5 forecasting competition due to their strong performance and ease of use \cite{makridakis_m5_2022,januschowski_forecasting_2022}. Moreover, our e-commerce partner's primary product forecasting is a LightGBM-based model, so we expect results from off\-line experiments on public datasets to transfer to our proprietary setting when using the same base forecasting model. We compare the performance of our LightGBM models against traditional statistical methods ARIMA \cite{box_distribution_1970}, ETS \cite{hyndman_forecasting_2008}, Theta \cite{assimakopoulos_theta_2000}, SeasonalNaive \cite{hyndman_forecasting_2021}, Naive \cite{hyndman_forecasting_2021} and Croston \cite{croston_forecasting_1972}. 
  We note that deep learning based approaches are becoming more prevalent in e-commerce \cite{kunz_deep_2023}, especially with the rise of the Transformer-architecture in forecasting models \cite{lim_temporal_2021,li_enhancing_2019}. We consider this for future work, and did not consider this for our study as (i) the cloud cost to operate these models is 10x higher for our e-commerce partner than a tree-based model, and (ii) none of the neural-network based methods are able to scale to the size of our e-commerce partner, as explained in Section~\ref{sec:scalingissuesneuralmethods}. 

  \paragraph{Experimental setup} To test our hierarchical sparse loss function against baseline forecasting systems, we consider the following scenarios:
  \begin{enumerate}
    \item \texttt{Bottom-up}. We train a single global model only on the bottom-level time series. \rev{Subsequently, the bottom-level forecasts are aggregated to obtain the aggregated (reconciled) forecasts.}
    \item \texttt{Separate aggregations}. We train separate mo\-dels for every aggregation in the hierarchy, resulting in 12 models for the entire M5 dataset.
    \item \texttt{Global}. We train a single global model on all time series in the dataset, including all the aggregations.
  \end{enumerate}
  For the first scenario in our experiments (\texttt{Bottom-up}), we vary both the \textit{objective} (i.e. loss function that is optimized by LightGBM) and the \textit{evaluation metric} (i.e. the loss function that governs early-stopping during hyperparameter optimization). For the \textit{objective}, we consider the \textit{squared error loss} (SL), the \textit{Tweedie loss} (TL) and our \textit{sparse hierarchical loss} (HL). The Tweedie loss is a loss function that assumes that the time series follow a distribution somewhere in between a Poisson and a Gamma distribution, which is useful in zero-inflated settings such as retail demand forecasting. It is a loss function that was favored by contestants in the M5 forecasting competition \cite{januschowski_forecasting_2022}, and it is the loss also used in the primary forecasting system of our e-commerce partner.

  For the latter two scenarios, we will obtain non-coherent forecasts. Thus, these methods require a reconciliation post-processing step to reconcile the forecasts to the hierarchy. We employ the following cross-sectional reconciliation methods:
  \begin{itemize}
    \item \textbf{Base}. No reconciliation is performed.
    \item \textbf{OLS}. Ordinary Least Squares (OLS) \cite{hyndman_optimal_2011}, where \(W\) in Eq.~\eqref{eq:p1} is the identity matrix.
    \item \textbf{WLS-struct} and \textbf{WLS-var}. Weighted Least Squares (WLS) \cite{wickramasuriya_optimal_2019}, where \(W\) in Eq.~\eqref{eq:p1} is a diagonal matrix containing respectively the sum of the rows of \(S\) (\textit{WLS-struct}) or the in-sample forecast errors (\textit{WLS-var}).
    \item \textbf{MinT-shrink}. \textit{Trace Minimization} \cite{wickramasuriya_optimal_2019}, where \(W\) in Eq.~\eqref{eq:p1} is the shrunk covariance matrix of in-sample forecast errors. We also experimented with using the non-shrunk covariance matrix of the in-sample forecast errors (\textit{MinT-cov}), but this produced erroneous/high variance results, which we attribute to precisely the motivation to shrink the covariance matrix in \textit{MinT-shrink}: to reduce the variance when the amount of time series considered becomes very large.
    \item \textbf{ERM}. The \textit{Empirical Risk Minimization} (ERM) method \cite{bentaieb_regularized_2019}. Due to computational issues explained in Section~\ref{sec:scalingissuesreconmethods}, we were not able to apply the regularized ERM variants to our experiments, but only the unregularized variant.
  \end{itemize}

\noindent%
We optimize the hyperparameters of each of the LightGBM models by Bayesian hyperparameter optimization using Optuna \cite{akiba_optuna_2019}. The settings for the hyperparameter optimization can be found in \ref{app:hyperparam}. Each model is trained for 10 different random seeds, and our results are based on the mean and standard deviation of those 10 rollouts. \rev{For the traditional statistical methods we use Nixtla's StatsForecast \cite{garza_statsforecast_2022}, which includes automatic optimization of the hyperparameters of the statistical methods.}

  \paragraph{Evaluation} We evaluate our results for every aggregation in the hierarchy using the Root Mean Squared Error (RMSE) and Mean Absolute Error (MAE) \cite{hyndman_forecasting_2021}. In the results section, we present the RMSE / MAE relative to the \texttt{Bottom-up} scenario using the squared-loss objective with the squared-loss metric. For absolute values and standard deviation of the results, see \ref{app:experiments}.

  \paragraph{Results -- LightGBM as baseline model} For our first experiment, we only consider cross-sectional hierarchies (i.e., \rev{\(S^{te} = I_{n_b^{te}}\)}). We present our results on relative RMSE using LightGBM as baseline model in Table~\ref{tab:allstores_rel_rmse} and conclude the following:
  \begin{itemize}
    \item The best method is the \texttt{Bottom-up}-scenario combined with our sparse hierarchical loss as objective, outperforming the baseline by 0--20\% across aggregations. This holds for both settings in which we use our sparse hierarchical loss.
    \item Even when we only use our sparse hierarchical loss as an evaluation metric during training whilst optimizing the standard squared loss (the SL/HL scenario), we already see a small improvement of $\pm$5\% across aggregations. 
    \item Even though the Tweedie loss improves over the baseline loss, our sparse hierarchical loss function still outperforms it by $\pm$5\% across aggregations.
    \item From the reconciliation methods, \textit{MinT-shrink} and \textit{WLS-var} perform best in the \texttt{Separate aggregations}-scenario, although the performance delta across aggregations is still $\pm$5-30\% as compared to the best (our) method.
  \end{itemize}
  For relative MAE, we present our results in Table~\ref{tab:allstores_rel_mae}. We find that overall, our sparse hierachical loss still performs best by $\pm$5\% compared to other loss functions and scenarios. However, the results are more nuanced: we find that \textit{MinT-shrink} in the \texttt{Separate aggregations}-scenario performs strong as well. In addition, we also find that the Tweedie loss (TL) performs relatively well. This finding corroborates the usefullness of the TL in intermittent demand settings, such as retail, where zero demand is often observed.

  Next, we compare our findings against the forecasting results when employing different baseline models in Table~\ref{tab:otherbaselinemodels}. For brevity, we only show the metrics for all time series combined (incl. aggregations). In addition, we only show a single reconciliation method for the other baseline models, as we found little difference in results when employing different reconciliation methods. We then find that in terms of RMSE, our sparse hierachical loss outperforms the other baseline models by at least 50\%, and in terms of MAE by at least 10\%. This verifies that on this dataset and with this type of problem, using a more complex model such as LightGBM greatly improves forecasting performance, as was also shown in the M5 forecasting competition \cite{makridakis_m5_2022}.

  \begin{table}[t]
    \caption{Forecasting results for all time series (incl. aggregations) on the M5 dataset, using different baseline models. We show absolute and relative RMSE and MAE. Lower is better, and bold indicates the best performing method.}
    \label{tab:otherbaselinemodels}
    \centering
    {\small\setlength{\tabcolsep}{2pt} 
    \rev{\begin{tabular}{l l cccc}
    \toprule 
     & &\multicolumn{2}{ c }{RMSE}   &\multicolumn{2}{ c }{MAE} \\
     \cmidrule(r){3-4} \cmidrule(r){5-6}
    Model &Reconciliation &Abs. &Rel. &Abs. &Rel. \\
    \midrule																	
    LightGBM (SL/SL)	&None	&\textit{22.39}	&\textit{1.00}  &\textit{2.20}	&\textit{1.00}	\\
    LightGBM (HL/HL)	&None	&\textbf{19.54}	&\textbf{0.87}  &\textbf{2.10}	&\textbf{0.95}	\\
    LightGBM (HL/SL)	&None	&\textbf{19.59}	&\textbf{0.88}  &\textbf{2.10}	&\textbf{0.95}	\\
    ARIMA	&MinT-shrink	&39.88	&2.43	&1.78	&1.10	\\
    ETS	&MinT-shrink	&36.48	&2.35	&1.63	&1.07	\\
    Theta	&MinT-shrink	&36.66	&2.39	&1.64	&1.08	\\
    Croston	&None	&39.40	&2.76	&1.76	&1.25	\\
    Naive	&None	&74.91	&3.95	&3.35	&1.80	\\
    Seasonal Naive	&None	&39.40	&2.76	&1.76	&1.25	\\        
    \bottomrule
    \end{tabular}}}
    \end{table}

    \begin{table}[t]
      \caption{Forecasting results for all time series (incl. aggregations) on the M5 dataset, ablating for the use of cross-sectional and temporal hierarchies. We show absolute and relative RMSE and MAE, with the standard deviation in brackets. Lower is better, and bold indicates the best performing method. Note that when not using cross-sectional nor temporal aggregations, the hierarchical loss is equal to the standard squared error loss.}
      \label{tab:ablation_hierarchies}
      \centering
      {\small\setlength{\tabcolsep}{2pt} 
      \rev{\begin{tabular}{l l cccc}
      \toprule 
      \multicolumn{2}{ c }{Hierarchies} &\multicolumn{2}{ c }{RMSE}   &\multicolumn{2}{ c }{MAE} \\
      \cmidrule(r){1-2}  \cmidrule(r){3-4} \cmidrule(r){5-6}
      Cross- & \\
      sectional &Temporal &Abs. &Rel. &Abs. &Rel. \\
      \midrule																	
      No	&No	&\textit{22.39} (0.16)	&\textit{1.00}	&\textit{2.20} (0.01)	&\textit{1.00}	\\
      Yes	&No	&\textbf{19.54}	(0.38) &\textbf{0.87}	&\textbf{2.10} (0.01)	&\textbf{0.95}	\\
      Yes	&Yes	&29.81 (1.52)	&1.33	&2.47	(0.04)  &1.12	\\
      No	&Yes	&26.65 (0.32)	&1.19	&2.36	(0.01) &1.07	\\
      Random	&No	&\revv{23.54 (0.73)}	&\revv{1.05}	&\revv{2.19	(0.02)} &\revv{1.00}	\\
      \bottomrule
      \end{tabular}}}
      \end{table}

      \begin{table}[t]
        \caption{Forecasting results for all time series (incl. aggregations) on the M5 dataset, ablating for the use of cross-sectional and temporal hierarchies. We show relative RMSE for several forecasting day buckets of the forecast. Lower is better, and bold indicates the best performing method.}
        \label{tab:ablation_hierarchies_timesteps}
        \centering
        {\small\setlength{\tabcolsep}{2pt} 
        \begin{tabular}{l l  cccc}
        \toprule 
        \multicolumn{2}{ c }{Hierarchies} &\multicolumn{4}{ c }{Forecast day}  \\
        \cmidrule(r){1-2}  \cmidrule(r){3-6}
        Cross-sectional &Temporal &1--7 &8--14 &15--21 &22--28 \\
        \midrule																	
        No	&No	&\textit{1.00}	&\textit{1.00}	&\textit{1.00}	&\textit{1.00}	\\
        Yes	&No	&\textbf{0.98}	&0.99	&\textbf{0.81}	&\textbf{0.80}	\\
        Yes	&Yes	&1.75	&1.50	&0.96	&1.66	\\
        No	&Yes	&1.37	&1.24	&0.97	&1.41	\\
        Random	&No	&\revv{1.31}	&\textbf{\revv{0.94}}	&\revv{1.07}	&\revv{0.87}	\\        
        \bottomrule
        \end{tabular}}
        \end{table}

  \begin{sidewaystable*}[t]
    \caption{Forecasting results for all stores on the M5 dataset, using LightGBM as baseline model. We report relative RMSE as compared to the baseline (shown in italic). Lower is better, and bold indicates best method for the aggregation, taking into account standard deviation of the best method across the 10 seeds. For absolute values and standard deviation of the results, see \ref{app:experiments}. \rev{The Bottom-up scenario using the HL loss commonly outperforms all other scenarios.}}
    \label{tab:allstores_rel_rmse}
    \centering
    {\small\setlength{\tabcolsep}{2pt} 
    \begin{tabular}{l l l  rrrrrrrrrrrrr}
    \toprule 
     &&&& &  &\multicolumn{3}{ c }{Store}   &\multicolumn{2}{ c }{Product} &\multicolumn{3}{ c }{State} \\
     \cmidrule(r){7-9} \cmidrule(r){10-11} \cmidrule(r){12-14}
    Scenario/Objective & Metric  & Reconciliation &Product	&Department	&Category &Department	&Category	&Total &Store	&State &Department &Category &Total	&Total	&All series \\
    \midrule																	
    \texttt{Bottom-up}																	\\
    \hspace{0.1cm} 	\textit{SL}	&\textit{SL}	&\textit{None}	&\textbf{\textit{1.00}}	&\textit{1.00}&\textit{1.00}&\textit{1.00}&\textit{1.00}&\textit{1.00}&\textit{1.00}&\textit{1.00}&\textit{1.00}&\textit{1.00}&\textit{1.00}&\textit{1.00}&\textit{1.00}	\\
    \hspace{0.1cm} 	SL	&HL	&None	&\textbf{1.00}	&0.97	&0.95	&0.98	&0.97	&0.98	&0.99	&1.00	&0.98	&0.96	&0.99	&1.00	&0.98	\\
    \hspace{0.1cm} 	HL	&HL	&None	&\textbf{1.00}	&\textbf{0.88}	&\textbf{0.80}	&\textbf{0.93}	&\textbf{0.89}	&\textbf{0.93}	&\textbf{0.97}	&\textbf{0.99}	&\textbf{0.89}	&\textbf{0.84}	&\textbf{0.88}	&\textbf{0.87}	&\textbf{0.87}	\\
    \hspace{0.1cm} 	HL	&SL	&None	&\textbf{1.00}	&\textbf{0.88}	&\textbf{0.81}	&\textbf{0.94}	&\textbf{0.90}	&\textbf{0.94}	&\textbf{0.96}	&\textbf{0.98}	&\textbf{0.89}	&\textbf{0.84}	&\textbf{0.88}	&\textbf{0.87}	&\textbf{0.88}	\\
    \hspace{0.1cm} 	TL	&HL	&None	&\textbf{1.00}	&0.95	&0.96	&0.99	&1.00	&1.00	&0.99	&1.00	&0.97	&0.98	&0.98	&0.96	&0.97	\\
    \hspace{0.1cm} 	TL	&SL	&None	&\textbf{1.00}	&0.94	&0.93	&1.00	&1.00	&1.00	&0.99	&1.00	&0.97	&0.98	&0.98	&0.93	&0.96	\\
    \hspace{0.1cm} 	TL	&TL	&None	&1.17	&2.72	&2.81	&1.76	&1.83	&1.73	&1.52	&1.33	&2.15	&2.18	&2.08	&2.71	&2.38	\\
    \midrule																	
    \texttt{Sep. agg.}																	\\
    \hspace{0.1cm} 	SL	&SL	&Base	&\textbf{1.00}	&1.44	&1.29	&1.19	&1.14	&1.14	&1.01	&0.99	&1.23	&1.34	&1.27	&1.60	&1.35	\\
    \hspace{0.1cm} 	SL	&SL	&OLS	&\textbf{1.00}	&1.39	&1.41	&1.10	&1.06	&1.07	&1.00	&1.00	&1.20	&1.19	&1.23	&1.50	&1.30	\\
    \hspace{0.1cm} 	SL	&SL	&WLS-struct	&\textbf{1.00}	&1.26	&1.37	&1.03	&1.05	&1.02	&0.99	&0.99	&1.11	&1.16	&1.16	&1.39	&1.23	\\
    \hspace{0.1cm} 	SL	&SL	&WLS-var	&\textbf{1.00}	&1.12	&1.23	&0.99	&1.02	&0.99	&0.99	&0.99	&1.03	&1.09	&1.07	&1.22	&1.12	\\
    \hspace{0.1cm} 	SL	&SL	&MinT-shrink	&\textbf{1.00}	&1.15	&1.27	&0.97	&0.99	&0.97	&1.00	&1.00	&1.03	&1.09	&1.09	&1.30	&1.15	\\
    \hspace{0.1cm} 	SL	&SL	&ERM	&1.22	&1.25	&1.29	&1.07	&1.03	&1.07	&1.17	&1.22	&1.17	&1.14	&1.22	&1.49	&1.26	\\
    \midrule																	
    \texttt{Global}																	\\
    \hspace{0.1cm} 	SL	&SL	&Base	&1.02	&1.33	&1.45	&1.09	&1.10	&1.10	&1.03	&1.03	&1.25	&1.27	&1.81	&1.57	&1.46	\\
    \hspace{0.1cm} 	SL	&SL	&OLS	&1.01	&1.32	&1.39	&1.07	&1.09	&1.16	&1.02	&1.02	&1.20	&1.25	&1.38	&1.49	&1.34	\\
    \hspace{0.1cm} 	SL	&SL	&WLS-struct	&1.01	&1.38	&1.54	&1.08	&1.13	&1.11	&1.03	&1.02	&1.19	&1.28	&1.27	&1.55	&1.36	\\
    \hspace{0.1cm} 	SL	&SL	&WLS-var	&1.01	&1.51	&1.70	&1.18	&1.27	&1.22	&1.03	&1.02	&1.31	&1.43	&1.38	&1.66	&1.48	\\
    \hspace{0.1cm} 	SL	&SL	&MinT-shrink	&1.03	&1.26	&1.41	&1.05	&1.10	&1.15	&1.06	&1.05	&1.11	&1.17	&1.24	&1.54	&1.30	\\
    \hspace{0.1cm} 	SL	&SL	&ERM	&1.21	&1.59	&1.69	&1.26	&1.28	&1.34	&1.20	&1.23	&1.45	&1.49	&1.61	&1.80	&1.59	\\      
    \bottomrule
    \end{tabular}}
    \end{sidewaystable*}

    \begin{sidewaystable*}[t]
      \caption{Forecasting results for all stores on the M5 dataset, using LightGBM as baseline model. We report relative MAE as compared to the baseline (shown in italic). Lower is better, and bold indicates best method for the aggregation, taking into account standard deviation of the best method across the 10 seeds. For absolute values and standard deviation of the results, see \ref{app:experiments}. \rev{The \texttt{Bottom-up} scenario using the HL loss commonly outperforms all other scenarios.}}
      \label{tab:allstores_rel_mae}
      \centering
      {\small\setlength{\tabcolsep}{2pt} 
      \begin{tabular}{l l l rrrrrrrrrrrrr}
      \toprule 
       &&&& &  &\multicolumn{3}{ c }{Store}   &\multicolumn{2}{ c }{Product} &\multicolumn{3}{ c }{State} \\
       \cmidrule(r){7-9} \cmidrule(r){10-11} \cmidrule(r){12-14}
      Scenario/Objective & Metric  & Reconciliation &Product	&Department	&Category &Department	&Category	&Total &Store	&State &Department &Category &Total	&Total	&All series \\
      \midrule																	
      \texttt{Bottom-up}																	\\
      \hspace{0.1cm} 	\textit{SL}	&\textit{SL}	&\textit{None}	&\textit{1.00}	&\textit{1.00}&\textit{1.00}&\textit{1.00}&\textit{1.00}&\textit{1.00}&\textit{1.00}&\textit{1.00}&\textit{1.00}&\textit{1.00}&\textit{1.00}&\textit{1.00}&\textit{1.00}	\\
      \hspace{0.1cm} 	SL	&HL	&None	&1.00	&0.98	&0.96	&0.98	&0.98	&0.99	&1.00	&1.00	&0.97	&0.97	&1.00	&1.02	&0.99	\\
      \hspace{0.1cm} 	HL	&HL	&None	&0.99	&\textbf{0.81}	&\textbf{0.78}	&\textbf{0.90}	&\textbf{0.89}	&0.94	&0.98	&0.99	&\textbf{0.85}	&\textbf{0.83}	&\textbf{0.89}	&\textbf{0.88}	&\textbf{0.95}	\\
      \hspace{0.1cm} 	HL	&SL	&None	&0.99	&\textbf{0.81}	&\textbf{0.77}	&\textbf{0.90}	&\textbf{0.90}	&0.94	&0.98	&0.99	&\textbf{0.85}	&\textbf{0.83}	&\textbf{0.89}	&\textbf{0.87}	&\textbf{0.95}	\\
      \hspace{0.1cm} 	TL	&HL	&None	&0.98	&\textbf{0.83}	&0.82	&0.93	&0.94	&0.96	&0.97	&0.98	&0.88	&0.89	&0.92	&\textbf{0.88}	&\textbf{0.95}	\\
      \hspace{0.1cm} 	TL	&SL	&None	&0.99	&0.85	&0.84	&0.95	&0.95	&0.98	&0.98	&0.99	&0.90	&0.91	&0.96	&\textbf{0.90}	&\textbf{0.96}	\\
      \hspace{0.1cm} 	TL	&TL	&None	&1.02	&2.06	&2.39	&1.47	&1.61	&1.65	&1.14	&1.07	&1.69	&1.88	&1.97	&2.86	&1.29	\\
      \midrule																	
      \texttt{Sep. agg.}																	\\
      \hspace{0.1cm} 	SL	&SL	&Base	&1.00	&1.11	&1.06	&1.02	&1.10	&1.08	&0.96	&0.97	&1.03	&1.15	&1.20	&1.55	&1.02	\\
      \hspace{0.1cm} 	SL	&SL	&OLS	&\textbf{0.96}	&1.08	&1.16	&1.01	&0.99	&1.00	&0.96	&0.97	&1.00	&1.01	&1.12	&1.49	&0.99	\\
      \hspace{0.1cm} 	SL	&SL	&WLS-struct	&0.97	&0.99	&1.13	&0.92	&0.95	&0.95	&\textbf{0.96}	&\textbf{0.97}	&0.93	&0.98	&1.04	&1.42	&0.98	\\
      \hspace{0.1cm} 	SL	&SL	&WLS-var	&0.98	&0.91	&1.00	&0.91	&0.92	&\textbf{0.92}	&0.96	&0.97	&0.90	&0.93	&0.95	&1.16	&\textbf{0.96}	\\
      \hspace{0.1cm} 	SL	&SL	&MinT-shrink	&0.97	&0.93	&1.05	&\textbf{0.90}	&\textbf{0.90}	&\textbf{0.91}	&0.96	&0.97	&0.89	&0.92	&0.98	&1.30	&\textbf{0.96}	\\
      \hspace{0.1cm} 	SL	&SL	&ERM	&1.18	&1.17	&1.21	&1.09	&1.06	&1.07	&1.19	&1.22	&1.11	&1.12	&1.19	&1.57	&1.18	\\
      \midrule																	
      \texttt{Global}																	\\
      \hspace{0.1cm} 	SL	&SL	&Base	&1.04	&1.04	&1.17	&1.00	&1.02	&1.05	&0.99	&0.99	&1.04	&1.09	&1.62	&1.58	&1.05	\\
      \hspace{0.1cm} 	SL	&SL	&OLS	&0.98	&1.09	&1.17	&1.05	&1.07	&1.13	&0.99	&1.00	&1.09	&1.13	&1.27	&1.50	&1.03	\\
      \hspace{0.1cm} 	SL	&SL	&WLS-struct	&0.99	&1.09	&1.26	&0.98	&1.01	&1.04	&1.00	&0.99	&1.00	&1.07	&1.15	&1.57	&1.02	\\
      \hspace{0.1cm} 	SL	&SL	&WLS-var	&0.99	&1.24	&1.40	&1.09	&1.13	&1.13	&1.01	&1.00	&1.14	&1.21	&1.25	&1.69	&1.06	\\
      \hspace{0.1cm} 	SL	&SL	&MinT-shrink	&0.99	&1.13	&1.24	&1.06	&1.08	&1.15	&1.02	&1.01	&1.05	&1.07	&1.15	&1.60	&1.04	\\
      \hspace{0.1cm} 	SL	&SL	&ERM	&1.17	&1.49	&1.62	&1.30	&1.34	&1.37	&1.19	&1.20	&1.39	&1.49	&1.58	&1.96	&1.27	\\
      \bottomrule
      \end{tabular}}
      \end{sidewaystable*}
          
  \paragraph{Analysis: impact of hierarchy} We investigate the impact of the choice of hierarchy. 
  
  \textit{Temporal hierarchies}. As we noted before, we only used cross-sectional aggregations in our first experiments. We now also include temporal aggregations by aggregating our bottom-level time series across years, weeks and months. We ablate for every setting and show the results in Table~\ref{tab:ablation_hierarchies}. Interestingly, we find that using temporal hierarchies jointly with cross-sectional hierarchies reduces forecasting performance by $\pm$35\% (RMSE) and $\pm$17\% (MAE). This setting is even worse than only using temporal hierarchies, which performs worse than using only cross-sectional hierarchies by $\pm$26\% (RMSE) and $\pm$12\% (MAE). We further analyze these results by studying the RMSE across the forecast days in Table~\ref{tab:ablation_hierarchies_timesteps}. As noted before, we forecast 28 days ahead, and each forecast is created by recursively applying the one-step ahead LightGBM model. We find that as we forecast further into the future, the setting with only using cross-sectional aggregations starts to perform better by up to 20\% as compared to the baseline where we do not use any aggregations. Again, the setting where we employ temporal hierarchies too shows relatively bad performance across all forecast day buckets.  
  
  \textit{Random hierarchies}. In hierarchical forecasting pro\-blems, the aggregation matrices \(S^{cs}\) and \(S^{te}\) are commonly fixed a priori and considered constant during training and prediction. \revv{As we are performing the reconciliation in an end-to-end fashion during training, we can modify these matrices. This allows us to understand the robustness of our solution to possible misspecification of the hierarchy, and more generally, to what extent the choice of the hierarchy has an effect on forecasting performance. We perform an experiment by randomly sampling an \(S^{cs}\)-matrix before we start the LightGBM training process. We sample uniformly at random (i) a number of levels for the cross-sectional hierarchy and (ii) a number of maximum categories for the level and construct a random \(S^{cs}\)-matrix to be used in the gradient \eqref{eq:hfpbu_grad} and \rev{second-order derivative} \eqref{eq:hfpbu_hess}. We validate and test on the `true' \(S^{cs}\)-matrix. We present the results in Table~\ref{tab:ablation_hierarchies} and Table~\ref{tab:ablation_hierarchies_timesteps}, under `Random'. In Table~\ref{tab:ablation_hierarchies}, we find that on RMSE, forecasting performance deteriorates by about 5\% as compared to the baseline using no hierarchies, and by about 20\% as compared to the best setting in which we use the correct cross-sectional hierarchies during training. In Table~\ref{tab:ablation_hierarchies_timesteps}, we find that `Random' performs poorly on the first forecast period, whereas it performs strong on the second and final week of the forecast period. Thus, misspecification of the hierarchy can severely deteriorate forecasting performance, but the relatively strong performance at some forecast intervals in Table~\ref{tab:ablation_hierarchies_timesteps} could also indicate that a better hierarchy randomization strategy might lead to improved forecast results. We leave this for future work.}
  
  \paragraph{Analysis: time complexity} \label{par:timecomplexity} We investigate the computational time complexity required to perform training and prediction for each scenario and present the results in Table~\ref{tab:complexity}. The training and prediction time complexity is indicated by how respectively the training time and prediction time scales with respect to the default LightGBM training and prediction time complexity. We first investigate the case where we only consider cross-sectional hierarchies. This case is indicated by `HL' in Table~\ref{tab:complexity}. First, we note that adding our hierarchical loss objective adds a component to the time complexity that scales with \rev{\(\left(n_b^{cs}\right)^3\)}, as we need to compute \eqref{eq:hfp_grad}. However, our sparse implementation of the hierarchical loss reduces this component from \rev{\(\left(n_b^{cs}\right)^3\) to \(\left(n_b^{cs}\right)^2l^{cs}\)}, effectively reducing the scaling from cubic to quadratic in the number of bottom-level time series, as \rev{\(l^{cs}\)} is generally small. In the reconciliation scenarios, we always need to compute a matrix inversion to solve Eq.~\eqref{eq:p1} that scales cubically with the number of cross-sectional aggregations \rev{\(n_a^{cs}\)} or with the total number of time series \rev{\(n^{cs}\)}. The first is not problematic as generally \rev{\(n_a^{cs} \ll n_b^{cs}\)} in large-scale settings, but methods with this time complexity consequently trade in performance, as we observed in Table~\ref{tab:allstores_rel_rmse}. To empirically verify the differences in asymptotic time complexity, we recorded the training and prediction time for each scenario. We show timings for training and prediction for a single store of the M5 dataset (4M training samples) and for the entire M5 dataset (52M training samples), to provide an indication of scaling when the problem size increases by an order of magnitude. First, we note that using our sparse implementation of the HL reduces training time by a factor of \(3\times\) when training for all stores. Second, our sparse HL has a prediction time similar to the baseline (SL). 
  Next, we find that the training time of our sparse hierarchical loss is two orders of magnitude faster than reconciliation methods in the \texttt{Separate aggregations}-scenario. This is mainly due to the many individual models that need to be trained in this scenario and thus shows a clear benefit of having just a single model. We observe an order of magnitude difference in prediction time when comparing the sparse hierarchical loss to the \texttt{Separate aggregations}-scenario when predictinf all stores. Again, this shows a clear benefit of having just a single model for this forecasting task. 
  For the \texttt{Global}-scenario, we see that reconciliation methods require a smaller training time when training for all stores (about twice less), however that scenario also did not give strong forecasting performance as we established in Table~\ref{tab:allstores_rel_rmse}. Also, the prediction time using our sparse HL is an order of magnitude lower. As ML costs in production systems mainly consists of prediction costs, having a lower prediction time is beneficial.\footnote{For example, Google designed its first TPU for inference: \href{https://techcrunch.com/2017/05/17/google-announces-second-generation-of-tensor-processing-unit-chips}{\texttt{https://techcrunch.com/2017/05/17/google-announces- second-generation-of-tensor-processing-unit-chips}}.}
  Finally, we also show the time complexity of using both cross-sectional and temporal hierarchies jointly, as indicated by `\(HL+\)' in Table~\ref{tab:complexity}. Adding temporal hierarchies adds another matrix multiplication that scales with the number of timesteps to the complexity. In our experiments, we find that adding temporal hierarchies results in a twice higher training time when training for all stores and a 50\% higher prediction time when predicting for all stores. We view it as potential future work to investigate how to perform this end-to-end learning of both cross-sectional and temporal hierarchies even more efficiently. 

  To conclude, we showed that our sparse HL incurs some additional training overhead but no additional prediction overhead as compared to the base case SL, whereas it does not require the additional reconciliation step that reconciliation methods require.  

  \begin{table*}[t]
    \rev{
    \caption{\rev{Computational time complexity and observed timings in seconds for all scenarios. The complexity is indicated by how respectively the training time and prediction time scales with respect to the default LightGBM training/prediction time \(L\), where \(n_b^{te}\) denotes the number of timesteps per time series, \(n_b^{cs}\) denotes the number of bottom-level time series in the hierarchy, \(n_b^{cs_l}\) the number of time series in each level in the hierarchy and \(l^{cs}\) the number of levels in the cross-sectional hierarchy, \(n^{cs}\) (\(n^{te}\)) the total number of cross-sectional (temporal) aggregations, and \(n_a = n - n_b\). \rev{The \texttt{Bottom-up} scenario using the HL loss is computationally more efficient than the \texttt{Separate aggregations} (both training and prediction) and \texttt{Global} (prediction) scenarios.}}}
    \label{tab:complexity}
    \centering\scalebox{0.8}{
    {\small\setlength{\tabcolsep}{1pt} 
    \begin{tabular}{l l lllrrrr}
    \toprule 
     &&  &\multicolumn{2}{ c }{Complexity}   &\multicolumn{2}{ c }{Training time (s)} &\multicolumn{2}{ c }{Prediction time (s)}  \\
     \cmidrule(r){4-5} \cmidrule(r){6-7} \cmidrule(r){8-9} 
    Scenario/Obj. &Metric  & Reconciliation &Training	&Prediction	&1 store &All stores &1 store &All stores \\
    \midrule						
    \texttt{Bottom-up}						\\
    \hspace{0.1cm} SL &SL &None & $O\left(L \left(n_b^{te}n_b^{cs}\right)\right)$ &$O\left(L \left(n_b^{te}n_b^{cs}\right) + n_b^{te}\left(n_b^{cs}\right)^3\right)$		&8	&173	&1.1	&11	\\
    \hspace{0.1cm} HL (dense) &HL (dense) &None & $O\left(L \left(n_b^{te}n_b^{cs} + n_b^{te}\left(n_b^{cs}\right)^3\right)\right)$ &$O\left(L \left(n^{te} n^{cs}\right) + n_b^{te}\left(n_b^{cs}\right)^3\right)$		&14	&1,185	&1.1	&10	\\
    \hspace{0.1cm} HL (sparse) &HL (sparse) &None & $O\left(L \left(n_b^{te}n_b^{cs} + n_b^{te}\left(n_b^{cs}\right)^2 l^{cs}\right)\right)$ &$O\left(L \left(n^{te} n^{cs}\right) + n_b^{te}\left(n_b^{cs}\right)^2 l^{cs}\right)$		&12	&318	&0.1	&11	\\
    \hspace{0.1cm} HL+ (sparse) &HL+ (sparse) &None & $O\left(L \left(n_b^{te}n_b^{cs} + n_b^{te}\left(n_b^{cs}\right)^2 l^{cs}n_b^{cs}\left(n_b^{te}\right)^2 l^{te}\right)\right)$ &$O\left(L \left(n^{te} n^{cs}\right) + n_b^{te}\left(n_b^{cs}\right)^2 l^{cs}n_b^{cs}\left(n_b^{te}\right)^2 l^{te}\right)$	&	&723	&	&15	\\
    \midrule 						
    \texttt{Sep. agg.} 						\\
    \hspace{0.1cm} SL &SL &Base & \multirow{6}{*}{$O\left(l^{cs} \cdot L\left(n_b^{cs_l} n_b^{te_l} \right)\right)$} &$O\left(l^{cs} \cdot L \left(n_b^{cs_l} n_b^{te_l}\right)\right)$ 		&\multirow{6}{*}{11}	&\multirow{6}{*}{36,018}	&4.4	&103	\\
    \hspace{0.1cm} SL &SL &OLS &  &$O\left(l^{cs} \cdot T\left(n_b^{cs_l} n_b^{te_l}\right) + \left(n_a^{cs}\right)^3\right)$  	&		&	&4.5	&149	\\
    \hspace{0.1cm} SL &SL &WLS-struct &  &$O\left(l^{cs} \cdot T\left(n_b^{cs_l} n_b^{te_l}\right) + \left(n_a^{cs}\right)^3\right)$ &			&	&4.5	&151	\\
    \hspace{0.1cm} SL &SL &WLS-var &  &$O\left(l^{cs} \cdot T\left(n_b^{cs_l} n_b^{te_l}\right) + \left(n_a^{cs}\right)^3\right)$		&	&	&4.5	&151	\\
    \hspace{0.1cm} SL &SL &MinT-shrink & &$O\left(l^{cs} \cdot T\left(n_b^{cs_l} n_b^{te_l}\right) + \left(n^{cs}\right)^3\right)$ &			&	&5.8	&305	\\
    \hspace{0.1cm} SL &SL &ERM &  &$O\left(l^{cs} \cdot T\left(n_b^{cs_l} n_b^{te_l}\right) + \left(n^{cs}\right)^3\right)$	&	&	&6.0	&239	\\
    \midrule 						
    \texttt{Global}  						\\
    \hspace{0.1cm} SL &SL &Base & \multirow{6}{*}{$O\left(L\left(n^{te}n^{cs}\right)\right)$} &$O\left(T\left(n^{te}n^{cs}\right)\right)$ 		&\multirow{6}{*}{4}	&\multirow{6}{*}{173}	&2.4	&71	\\
    \hspace{0.1cm} SL &SL &OLS &  &$O\left(T\left(n^{te}n^{cs}\right) + \left(n_a^{cs}\right)^3\right)$  		&	&	&2.5	&118	\\
    \hspace{0.1cm} SL &SL &WLS-struct &  &$O\left(T\left(n^{te}n^{cs}\right) + \left(n_a^{cs}\right)^3\right)$ 		&	&	&2.5	&120	\\
    \hspace{0.1cm} SL &SL &WLS-var &  &$O\left(T\left(n^{te}n^{cs}\right) + \left(n_a^{cs}\right)^3\right)$ 		&	&	&2.5	&120	\\
    \hspace{0.1cm} SL &SL &MinT-shrink &  &$O\left(T\left(n^{te}n^{cs}\right) + \left(n^{cs}\right)^3\right)$		&	&	&4.2	&274	\\
    \hspace{0.1cm} SL &SL &ERM &  &$O\left(T\left(n^{te}n^{cs}\right) + \left(n^{cs}\right)^3\right)$		&	&	&4.0	&207	\\
    \bottomrule				        
    \end{tabular}}}
    }
    \end{table*}
  
  \subsection{Proprietary datasets} \label{subsec:proprietarydatasets}
  At our e-commerce partner \textit{bol}, a LightGBM-based forecasting model is used as the primary product forecasting model. The model is used to forecast weekly product demand for 12 weeks. Every day, 12 separate models are trained, each tasked to forecast demand for a single week for every product. The model is used to forecast the majority of the products on sale at any moment, which are approximately 5 million unique items. We investigate the use of our sparse hierarchical loss function as a drop-in replacement for the existing Tweedie loss that is used within the company.

  \begin{table}
    \caption{\rev{Comparison of dataset characteristics between the M5 dataset and the proprietary dataset. We split the weekly demand by weekly demand buckets, and show the percentage of samples and percentage of demand for each bucket.}}
    \label{tab:dataset_compare}
    \rev{
    \centering
    \scalebox{0.9}{
    \begin{tabular}{l l l l l}
    \toprule 
    Weekly &\multicolumn{2}{ c }{\% samples} &\multicolumn{2}{ c }{\% demand} \\
    \cmidrule(r){2-3} \cmidrule(r){4-5}
    demand &M5 &Proprietary &M5 &Proprietary \\
    \toprule
    0	 &40.73\% 	 &34.15\% 	 &0\% 	 &0\% 	\\
    1	 &7.92\% 	 &19.23\% 	 &1.02\% 	 &3.99\% 	\\
    2--10	 &33.28\% 	 &37.31\% 	 &20.93\% 	 &31.6\% 	\\
    11-100	 &17.21\% 	 &8.89\% 	 &57.38\% 	 &46.01\% 	\\
    101--500	 &0.84\% 	 &0.39\% 	 &18.59\% 	 &14.27\% 	\\
    501+	 &0.02\% 	 &0.02\% 	 &2.08\% 	 &4.13\% 	\\     
    \midrule
    \texttt{Total} &100.00\% &100.00\% &100.00\% &100.00\% \\
    \bottomrule
  \end{tabular}}
  }
  \end{table}

  \begin{figure*}[!t] 
    \subfigure[RMSE]{\includegraphics[width=\columnwidth]{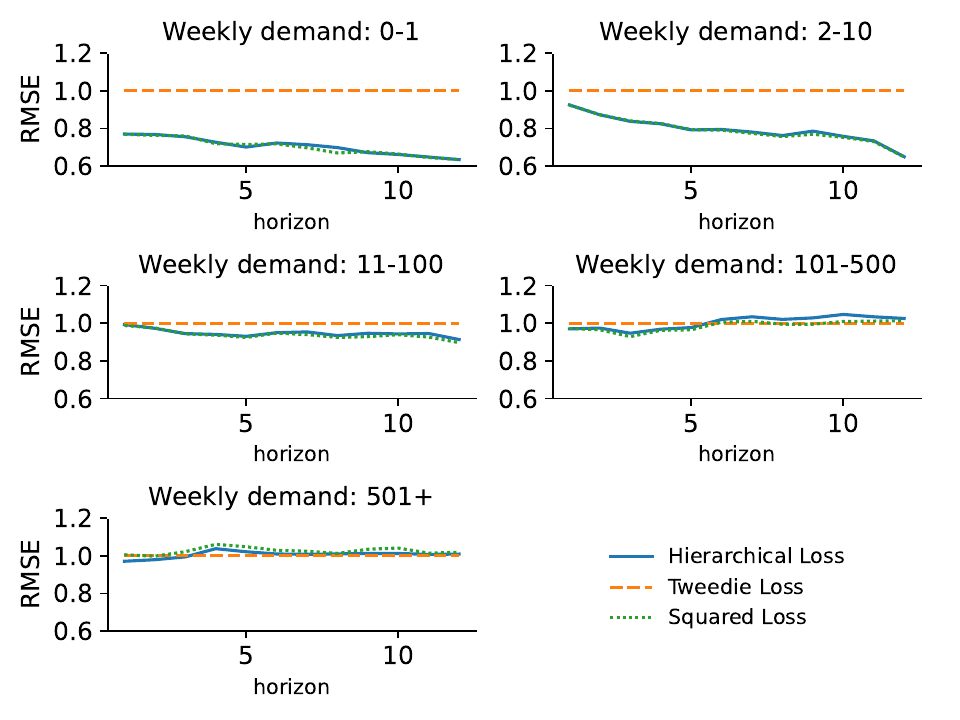}} \quad
    \subfigure[MAE]{\includegraphics[width=\columnwidth]{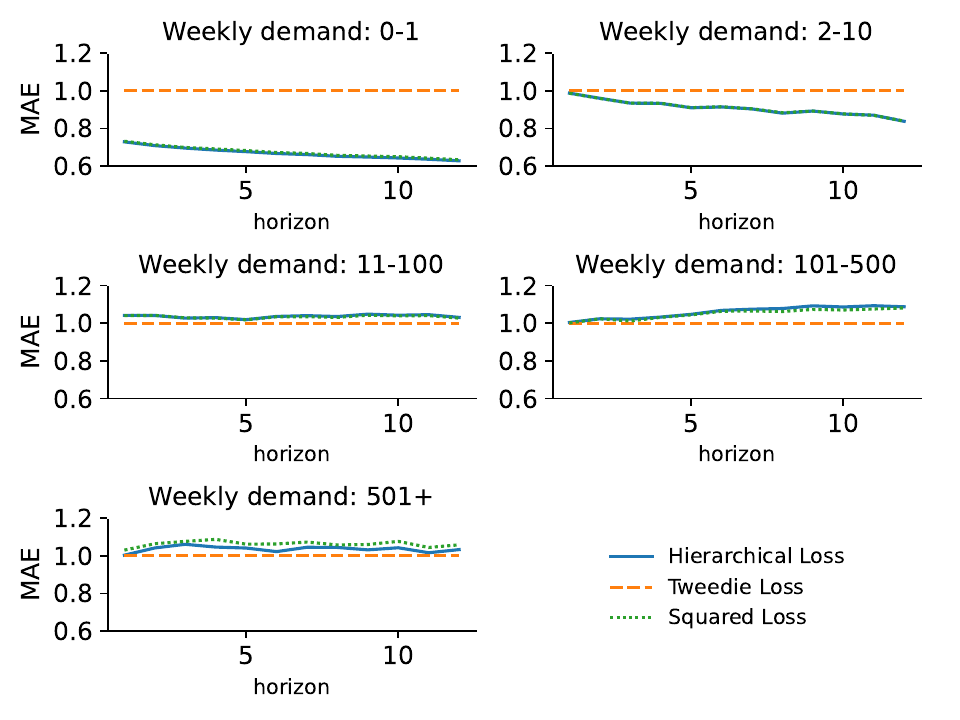}}
    \caption{Forecasting results for the primary product forecasting model at our e-commerce partner \textit{bol}. We show RMSE (a, left) and MAE (b, right) by weekly demand bucket relative to the Tweedie loss baseline for each forecasting horizon (week). The Hierarchical loss outperforms the Tweedie loss on RMSE and MAE on smaller weekly demand buckets.}
    \label{fig:bol_results}
  \end{figure*}

  \rev{\paragraph{Dataset} The off\-line dataset consists of 36M training samples from the period January 2017 to the end of June 2021. We test on 55M samples from the period July 2021 to January 2022. We show statistics of the proprietary dataset as compared to the M5 dataset in Table~\ref{tab:dataset_compare}, in which we split the weekly demand of both datasets according to weekly demand buckets used by our e-commerce partner. In Table~\ref{tab:dataset_compare}, we find that the M5 dataset and our proprietary dataset share demand characteristics in terms of sparsity (i.e., zero demand), which is 41\% for M5 and 34\% for our proprietary dataset, respectively. In general, we find that the two datasets share sufficient weekly demand density characteristics to warrant using our sparse HL on our proprietary dataset. The proprietary dataset contains 19 proprietary features, which are similar to those used in the M5 dataset (ref. Table~\ref{tab:features}), and consist of \begin{enumerate*}[label=(\roman*)] \item product categorical features, \item weekly demand (target) lagged features, and \item seasonality features. \end{enumerate*}}

  \paragraph{Experimental setup} \rev{The baseline model for every weekly forecast model is a LightGBM model with a \textit{Tweedie loss} (TL). We replace the TL with our HL and investigate forecasting performance on the test set. We apply log-scaling to the target values.} For the HL, we use the proprietary aggregations \textit{product\_group} and \textit{seasonality\_group}, each containing respectively $\pm$70 and $\pm$6,000 unique values. We have \rev{\(n_b^{cs} = \pm5\)M} bottom-level time series and \rev{\(n_a^{cs} = \pm6,070\)} aggregated time series across \rev{\(l^{cs} = 4\)} levels: \textit{product} (bottom-level), \textit{product\_group}, \textit{seasonality\_group} and \textit{total}.

\begin{figure}[t] 
    \includegraphics[width=\columnwidth, keepaspectratio]{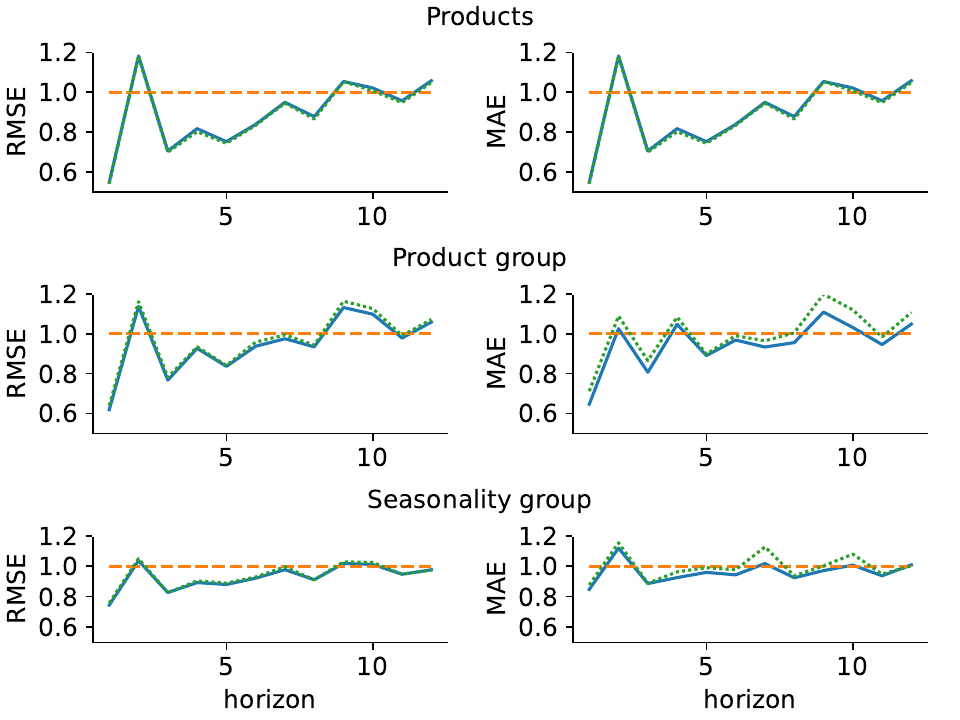}
    \caption{Forecasting results for the primary product forecasting model at our e-commerce partner \textit{bol}. We show RMSE (left column of figures) and MAE (right column of figures) by aggregation level relative to the Tweedie loss baseline for each forecasting horizon (week). The Hierarchical loss commonly outperforms the Tweedie loss on every aggregation level.}
    \label{fig:bol_level_results}
  \end{figure}

  \paragraph{Results} On average, we find that our sparse HL outperforms the existing TL model by about 1--2\% on RMSE and \(\pm\)10\% on MAE. We further investigate the performance by investigating how the RMSE and MAE vary across the 12 forecasting horizons and weekly demand buckets, and present the results in Figure~\ref{fig:bol_results}. We find that our sparse HL performs best on both RMSE and MAE on the lower weekly demand buckets (up to 100 products sold per week), where it outperforms the TL averaged over all the forecasting horizons. The TL is clearly better for higher weekly demand  buckets, commonly outperforming the HL and SL by up to 5\%. Next, we investigate forecasting performance across the cross-sectional hierarchies that we defined. We show the results in Figure~\ref{fig:bol_level_results}. We find that for most forecasting horizons, the HL and SL outperform the TL, with an average outperformance of the HL over the TL of \(\pm\)10\% at the  product level, \(\pm\)5\% at product group level and \(\pm\)4--7\% at seasonality group level. Hence, we are able to confirm some of the results we found in the M5 experiment, although the baseline SL performed quite strong in this experiment as well. We believe this is due to the M5 experiment having much more hierarchical levels (12 as compared to the 4 we used for our proprietary dataset experiment), since the HL is equal to the SL in the case of no hierarchies, and with fewer hierarchies the HL thus becomes closer to the SL. Hence, we believe our HL is most useful in settings with both many timeseries as well as many hierarchies.

  To conclude, we find that this experiment demonstrates the usefulness of our sparse HL applied to a production forecasting system at a major e-commerce platform.

\section{Conclusion} \label{sec:conclusion}
  We introduced a sparse hierarchical loss function to perform hierarchical forecasting in large-scale settings. We demonstrated that we are able to outperform existing hierarchical forecasting methods both in terms of performance as measured by RMSE and MAE by up to 10\% as well as in terms of computational time required to perform the end-to-end hierarchical forecasting in large-scale settings, reducing prediction time as compared to the best hierarchical forecasting reconciliation method by an order of magnitude. We empirically verified our sparse hierarchical loss in an offline test for \textit{bol}, where we confirmed the results from our offline test on the public M5 dataset. 

  In addition to our main contributions, one of our main learnings has been that we could not find a benefit of having multiple models for separate aggregations in the hierarchy, as the bottom-up scenario we employed consistently outperformed other scenarios. Secondly, we did not find a benefit of training a model whilst adhering to both cross-sectional and temporal hierarchies jointly. 

  Limitations of our work are that we did not consider the probabilistic forecasting setting, where reconciled forecasts are required across an entire forecast distribution.

  For future work, we aim to extend our work to the setting of probabilistic forecasting by combining our sparse hierarchical loss with existing probabilistic forecasting frameworks from, e.g., \cite{sprangers_probabilistic_2021, hasson_probabilistic_2021, stankeviciute_conformal_2021}. In addition, we seek to further investigate solutions for efficiently combining cross-sectional and temporal hierarchies. \revv{Finally, we aim to further understand the influence of hierarchy misspecification in the hierarchical forecasting setting.}

\section*{Acknowledgments}
This research was (partially) funded by the Hybrid Intelligence Center, a 10-year program funded by the Dutch Ministry of Education, Culture and Science through the Netherlands Organisation for Scientific Research\footnote{\url{https://www.hybrid-intelligence-centre.nl/}}, and Ahold Delhaize.\footnote{\url{https://www.aholddelhaize.com/}}

All content represents the opinion of the authors, which is not necessarily shared or endorsed by their respective employers and/or sponsors.

\bibliographystyle{elsarticle-num-names} 
\bibliography{lib}

\begin{thebibliography}{37}
\expandafter\ifx\csname natexlab\endcsname\relax\def\natexlab#1{#1}\fi
\providecommand{\url}[1]{\texttt{#1}}
\providecommand{\href}[2]{#2}
\providecommand{\path}[1]{#1}
\providecommand{\DOIprefix}{doi:}
\providecommand{\ArXivprefix}{arXiv:}
\providecommand{\URLprefix}{URL: }
\providecommand{\Pubmedprefix}{pmid:}
\providecommand{\doi}[1]{\href{http://dx.doi.org/#1}{\path{#1}}}
\providecommand{\Pubmed}[1]{\href{pmid:#1}{\path{#1}}}
\providecommand{\bibinfo}[2]{#2}
\ifx\xfnm\relax \def\xfnm[#1]{\unskip,\space#1}\fi
\bibitem[{Hyndman et~al.(2011)Hyndman, Ahmed, Athanasopoulos, and Shang}]{hyndman_optimal_2011}
\bibinfo{author}{R.~J. Hyndman}, \bibinfo{author}{R.~A. Ahmed}, \bibinfo{author}{G.~Athanasopoulos}, \bibinfo{author}{H.~L. Shang},
\newblock \bibinfo{title}{Optimal combination forecasts for hierarchical time series},
\newblock \bibinfo{journal}{Computational Statistics \& Data Analysis} \bibinfo{volume}{55} (\bibinfo{year}{2011}) \bibinfo{pages}{2579--2589}. \DOIprefix\doi{10.1016/j.csda.2011.03.006}.
\bibitem[{Athanasopoulos et~al.(2017)Athanasopoulos, Hyndman, Kourentzes, and Petropoulos}]{athanasopoulos_forecasting_2017}
\bibinfo{author}{G.~Athanasopoulos}, \bibinfo{author}{R.~J. Hyndman}, \bibinfo{author}{N.~Kourentzes}, \bibinfo{author}{F.~Petropoulos},
\newblock \bibinfo{title}{Forecasting with temporal hierarchies},
\newblock \bibinfo{journal}{European Journal of Operational Research} \bibinfo{volume}{262} (\bibinfo{year}{2017}) \bibinfo{pages}{60--74}. \DOIprefix\doi{10.1016/j.ejor.2017.02.046}.
\bibitem[{Rangapuram et~al.(2023)Rangapuram, Kapoor, Nirwan, Mercado, Januschowski, Wang, and {Bohlke-Schneider}}]{rangapuram_coherent_2023}
\bibinfo{author}{S.~S. Rangapuram}, \bibinfo{author}{S.~Kapoor}, \bibinfo{author}{R.~S. Nirwan}, \bibinfo{author}{P.~Mercado}, \bibinfo{author}{T.~Januschowski}, \bibinfo{author}{Y.~Wang}, \bibinfo{author}{M.~{Bohlke-Schneider}},
\newblock \bibinfo{title}{Coherent {{Probabilistic Forecasting}} of {{Temporal Hierarchies}}},
\newblock in: \bibinfo{booktitle}{Proceedings of {{The}} 26th {{International Conference}} on {{Artificial Intelligence}} and {{Statistics}}}, \bibinfo{publisher}{{PMLR}}, \bibinfo{year}{2023}, pp. \bibinfo{pages}{9362--9376}.
\bibitem[{Theodosiou and Kourentzes(2021)}]{theodosiou_forecasting_2021}
\bibinfo{author}{F.~Theodosiou}, \bibinfo{author}{N.~Kourentzes}, \bibinfo{title}{Forecasting with {{Deep Temporal Hierarchies}}}, \bibinfo{year}{2021}. \DOIprefix\doi{10.2139/ssrn.3918315}.
\bibitem[{Wickramasuriya et~al.(2019)Wickramasuriya, Athanasopoulos, and Hyndman}]{wickramasuriya_optimal_2019}
\bibinfo{author}{S.~L. Wickramasuriya}, \bibinfo{author}{G.~Athanasopoulos}, \bibinfo{author}{R.~J. Hyndman},
\newblock \bibinfo{title}{Optimal {{Forecast Reconciliation}} for {{Hierarchical}} and {{Grouped Time Series Through Trace Minimization}}},
\newblock \bibinfo{journal}{Journal of the American Statistical Association} \bibinfo{volume}{114} (\bibinfo{year}{2019}) \bibinfo{pages}{804--819}. \DOIprefix\doi{10.1080/01621459.2018.1448825}.
\bibitem[{Rangapuram et~al.(2021)Rangapuram, Werner, Benidis, Mercado, Gasthaus, and Januschowski}]{rangapuram_endtoend_2021}
\bibinfo{author}{S.~S. Rangapuram}, \bibinfo{author}{L.~D. Werner}, \bibinfo{author}{K.~Benidis}, \bibinfo{author}{P.~Mercado}, \bibinfo{author}{J.~Gasthaus}, \bibinfo{author}{T.~Januschowski},
\newblock \bibinfo{title}{End-to-{{End Learning}} of {{Coherent Probabilistic Forecasts}} for {{Hierarchical Time Series}}},
\newblock in: \bibinfo{booktitle}{Proceedings of the 38th {{International Conference}} on {{Machine Learning}}}, \bibinfo{publisher}{{PMLR}}, \bibinfo{year}{2021}, pp. \bibinfo{pages}{8832--8843}.
\bibitem[{Makridakis et~al.(2022)Makridakis, Spiliotis, and Assimakopoulos}]{makridakis_m5_2022}
\bibinfo{author}{S.~Makridakis}, \bibinfo{author}{E.~Spiliotis}, \bibinfo{author}{V.~Assimakopoulos},
\newblock \bibinfo{title}{M5 accuracy competition: {{Results}}, findings, and conclusions},
\newblock \bibinfo{journal}{International Journal of Forecasting} \bibinfo{volume}{38} (\bibinfo{year}{2022}) \bibinfo{pages}{1346--1364}. \DOIprefix\doi{10.1016/j.ijforecast.2021.11.013}.
\bibitem[{Ke et~al.(2017)Ke, Meng, Finley, Wang, Chen, Ma, Ye, and Liu}]{ke_lightgbm_2017}
\bibinfo{author}{G.~Ke}, \bibinfo{author}{Q.~Meng}, \bibinfo{author}{T.~Finley}, \bibinfo{author}{T.~Wang}, \bibinfo{author}{W.~Chen}, \bibinfo{author}{W.~Ma}, \bibinfo{author}{Q.~Ye}, \bibinfo{author}{T.-Y. Liu},
\newblock \bibinfo{title}{{{LightGBM}}: {{A Highly Efficient Gradient Boosting Decision Tree}}},
\newblock in: \bibinfo{booktitle}{Advances in {{Neural Information Processing Systems}} 30}, \bibinfo{publisher}{{Curran Associates, Inc.}}, \bibinfo{year}{2017}, pp. \bibinfo{pages}{3146--3154}.
\bibitem[{Januschowski et~al.(2022)Januschowski, Wang, Torkkola, Erkkil{\"a}, Hasson, and Gasthaus}]{januschowski_forecasting_2022}
\bibinfo{author}{T.~Januschowski}, \bibinfo{author}{Y.~Wang}, \bibinfo{author}{K.~Torkkola}, \bibinfo{author}{T.~Erkkil{\"a}}, \bibinfo{author}{H.~Hasson}, \bibinfo{author}{J.~Gasthaus},
\newblock \bibinfo{title}{Forecasting with trees},
\newblock \bibinfo{journal}{International Journal of Forecasting} \bibinfo{volume}{38} (\bibinfo{year}{2022}) \bibinfo{pages}{1473--1481}. \DOIprefix\doi{10.1016/j.ijforecast.2021.10.004}.
\bibitem[{B{\"o}se et~al.(2017)B{\"o}se, Flunkert, Gasthaus, Januschowski, Lange, Salinas, Schelter, Seeger, and Wang}]{bose_probabilistic_2017}
\bibinfo{author}{J.-H. B{\"o}se}, \bibinfo{author}{V.~Flunkert}, \bibinfo{author}{J.~Gasthaus}, \bibinfo{author}{T.~Januschowski}, \bibinfo{author}{D.~Lange}, \bibinfo{author}{D.~Salinas}, \bibinfo{author}{S.~Schelter}, \bibinfo{author}{M.~Seeger}, \bibinfo{author}{Y.~Wang},
\newblock \bibinfo{title}{Probabilistic demand forecasting at scale},
\newblock \bibinfo{journal}{Proceedings of the VLDB Endowment} \bibinfo{volume}{10} (\bibinfo{year}{2017}) \bibinfo{pages}{1694--1705}. \DOIprefix\doi{10.14778/3137765.3137775}.
\bibitem[{Benidis et~al.(2023)Benidis, Rangapuram, Flunkert, Wang, Maddix, Turkmen, Gasthaus, {Bohlke-Schneider}, Salinas, Stella, Aubet, Callot, and Januschowski}]{benidis_deep_2023}
\bibinfo{author}{K.~Benidis}, \bibinfo{author}{S.~S. Rangapuram}, \bibinfo{author}{V.~Flunkert}, \bibinfo{author}{Y.~Wang}, \bibinfo{author}{D.~Maddix}, \bibinfo{author}{C.~Turkmen}, \bibinfo{author}{J.~Gasthaus}, \bibinfo{author}{M.~{Bohlke-Schneider}}, \bibinfo{author}{D.~Salinas}, \bibinfo{author}{L.~Stella}, \bibinfo{author}{F.-X. Aubet}, \bibinfo{author}{L.~Callot}, \bibinfo{author}{T.~Januschowski},
\newblock \bibinfo{title}{Deep {{Learning}} for {{Time Series Forecasting}}: {{Tutorial}} and {{Literature Survey}}},
\newblock \bibinfo{journal}{ACM Computing Surveys} \bibinfo{volume}{55} (\bibinfo{year}{2023}) \bibinfo{pages}{1--36}. \DOIprefix\doi{10.1145/3533382}. \href{http://arxiv.org/abs/2004.10240}{{\tt arXiv:2004.10240}}.
\bibitem[{Chen and Guestrin(2016)}]{chen_xgboost_2016}
\bibinfo{author}{T.~Chen}, \bibinfo{author}{C.~Guestrin},
\newblock \bibinfo{title}{{{XGBoost}}: {{A Scalable Tree Boosting System}}},
\newblock in: \bibinfo{booktitle}{Proceedings of the 22nd {{ACM SIGKDD International Conference}} on {{Knowledge Discovery}} and {{Data Mining}}}, \bibinfo{publisher}{{ACM}}, \bibinfo{address}{{San Francisco California USA}}, \bibinfo{year}{2016}, pp. \bibinfo{pages}{785--794}. \DOIprefix\doi{10.1145/2939672.2939785}.
\bibitem[{Hyndman et~al.(2016)Hyndman, Lee, and Wang}]{hyndman_fast_2016}
\bibinfo{author}{R.~J. Hyndman}, \bibinfo{author}{A.~J. Lee}, \bibinfo{author}{E.~Wang},
\newblock \bibinfo{title}{Fast computation of reconciled forecasts for hierarchical and grouped time series},
\newblock \bibinfo{journal}{Computational Statistics \& Data Analysis} \bibinfo{volume}{97} (\bibinfo{year}{2016}) \bibinfo{pages}{16--32}. \DOIprefix\doi{10.1016/j.csda.2015.11.007}.
\bibitem[{Ben~Taieb et~al.(2017)Ben~Taieb, Taylor, and Hyndman}]{ben_taieb_coherent_2017}
\bibinfo{author}{S.~Ben~Taieb}, \bibinfo{author}{J.~W. Taylor}, \bibinfo{author}{R.~J. Hyndman},
\newblock \bibinfo{title}{Coherent {{Probabilistic Forecasts}} for {{Hierarchical Time Series}}},
\newblock in: \bibinfo{booktitle}{International {{Conference}} on {{Machine Learning}}}, \bibinfo{year}{2017}, pp. \bibinfo{pages}{3348--3357}.
\bibitem[{Ben~Taieb and Koo(2019)}]{bentaieb_regularized_2019}
\bibinfo{author}{S.~Ben~Taieb}, \bibinfo{author}{B.~Koo},
\newblock \bibinfo{title}{Regularized {{Regression}} for {{Hierarchical Forecasting Without Unbiasedness Conditions}}},
\newblock in: \bibinfo{booktitle}{Proceedings of the 25th {{ACM SIGKDD International Conference}} on {{Knowledge Discovery}} \& {{Data Mining}}}, \bibinfo{publisher}{{ACM}}, \bibinfo{address}{{Anchorage AK USA}}, \bibinfo{year}{2019}, pp. \bibinfo{pages}{1337--1347}. \DOIprefix\doi{10.1145/3292500.3330976}.
\bibitem[{Ben~Taieb(2017)}]{ben_taieb_sparse_2017}
\bibinfo{author}{S.~Ben~Taieb},
\newblock \bibinfo{title}{Sparse and {{Smooth Adjustments}} for {{Coherent Forecasts}} in {{Temporal Aggregation}} of {{Time Series}}},
\newblock in: \bibinfo{booktitle}{Proceedings of the {{Time Series Workshop}} at {{NIPS}} 2016}, \bibinfo{publisher}{{PMLR}}, \bibinfo{year}{2017}, pp. \bibinfo{pages}{16--26}.
\bibitem[{Athanasopoulos et~al.(2024)Athanasopoulos, Hyndman, Kourentzes, and Panagiotelis}]{athanasopoulos_forecast_2024}
\bibinfo{author}{G.~Athanasopoulos}, \bibinfo{author}{R.~J. Hyndman}, \bibinfo{author}{N.~Kourentzes}, \bibinfo{author}{A.~Panagiotelis},
\newblock \bibinfo{title}{Forecast reconciliation: {{A}} review},
\newblock \bibinfo{journal}{International Journal of Forecasting} \bibinfo{volume}{In Press} (\bibinfo{year}{2024}).
\bibitem[{Panagiotelis et~al.(2021)Panagiotelis, Athanasopoulos, Gamakumara, and Hyndman}]{panagiotelis_forecast_2021}
\bibinfo{author}{A.~Panagiotelis}, \bibinfo{author}{G.~Athanasopoulos}, \bibinfo{author}{P.~Gamakumara}, \bibinfo{author}{R.~J. Hyndman},
\newblock \bibinfo{title}{Forecast reconciliation: {{A}} geometric view with new insights on bias correction},
\newblock \bibinfo{journal}{International Journal of Forecasting} \bibinfo{volume}{37} (\bibinfo{year}{2021}) \bibinfo{pages}{343--359}. \DOIprefix\doi{10.1016/j.ijforecast.2020.06.004}.
\bibitem[{Girolimetto and Di~Fonzo(2023)}]{girolimetto_point_2023}
\bibinfo{author}{D.~Girolimetto}, \bibinfo{author}{T.~Di~Fonzo},
\newblock \bibinfo{title}{Point and probabilistic forecast reconciliation for general linearly constrained multiple time series},
\newblock \bibinfo{journal}{Statistical Methods \& Applications} \bibinfo{volume}{In Press} (\bibinfo{year}{2023}). \DOIprefix\doi{10.1007/s10260-023-00738-6}. \href{http://arxiv.org/abs/2305.05330}{{\tt arXiv:2305.05330}}.
\bibitem[{Han et~al.(2021)Han, Dasgupta, and Ghosh}]{han_simultaneously_2021}
\bibinfo{author}{X.~Han}, \bibinfo{author}{S.~Dasgupta}, \bibinfo{author}{J.~Ghosh},
\newblock \bibinfo{title}{Simultaneously {{Reconciled Quantile Forecasting}} of {{Hierarchically Related Time Series}}},
\newblock in: \bibinfo{booktitle}{Proceedings of {{The}} 24th {{International Conference}} on {{Artificial Intelligence}} and {{Statistics}}}, \bibinfo{publisher}{{PMLR}}, \bibinfo{year}{2021}, pp. \bibinfo{pages}{190--198}.
\bibitem[{Hyndman and Athanasopoulos(2021)}]{hyndman_forecasting_2021}
\bibinfo{author}{R.~J. Hyndman}, \bibinfo{author}{G.~Athanasopoulos}, \bibinfo{title}{Forecasting: {{Principles}} and {{Practice}} (3rd Ed)}, \bibinfo{publisher}{{OTexts: Melbourne}}, \bibinfo{address}{{Australia}}, \bibinfo{year}{2021}.
\bibitem[{Sch{\"a}fer and Strimmer(2005)}]{schafer_shrinkage_2005}
\bibinfo{author}{J.~Sch{\"a}fer}, \bibinfo{author}{K.~Strimmer},
\newblock \bibinfo{title}{A {{Shrinkage Approach}} to {{Large-Scale Covariance Matrix Estimation}} and {{Implications}} for {{Functional Genomics}}},
\newblock \bibinfo{journal}{Statistical Applications in Genetics and Molecular Biology} \bibinfo{volume}{4} (\bibinfo{year}{2005}). \DOIprefix\doi{10.2202/1544-6115.1175}.
\bibitem[{Touloumis(2015)}]{touloumis_nonparametric_2015}
\bibinfo{author}{A.~Touloumis},
\newblock \bibinfo{title}{Nonparametric {{Stein-type Shrinkage Covariance Matrix Estimators}} in {{High-Dimensional Settings}}},
\newblock \bibinfo{journal}{Computational Statistics \& Data Analysis} \bibinfo{volume}{83} (\bibinfo{year}{2015}) \bibinfo{pages}{251--261}. \DOIprefix\doi{10.1016/j.csda.2014.10.018}. \href{http://arxiv.org/abs/1410.4726}{{\tt arXiv:1410.4726}}.
\bibitem[{Makridakis et~al.(2021)Makridakis, Spiliotis, and Assimakopoulos}]{makridakis_m5_2021}
\bibinfo{author}{S.~Makridakis}, \bibinfo{author}{E.~Spiliotis}, \bibinfo{author}{V.~Assimakopoulos},
\newblock \bibinfo{title}{The {{M5}} competition: {{Background}}, organization, and implementation},
\newblock \bibinfo{journal}{International Journal of Forecasting}  (\bibinfo{year}{2021}). \DOIprefix\doi{10.1016/j.ijforecast.2021.07.007}.
\bibitem[{Virtanen et~al.(2020)Virtanen, Gommers, Oliphant, Haberland, Reddy, Cournapeau, Burovski, Peterson, Weckesser, Bright, Van Der~Walt, Brett, Wilson, Millman, Mayorov, Nelson, Jones, Kern, Larson, Carey, Polat, Feng, Moore, VanderPlas, Laxalde, Perktold, Cimrman, Henriksen, Quintero, Harris, Archibald, Ribeiro, Pedregosa, Van~Mulbregt, {SciPy 1.0 Contributors}, Vijaykumar, Bardelli, Rothberg, Hilboll, Kloeckner, Scopatz, Lee, Rokem, Woods, Fulton, Masson, H{\"a}ggstr{\"o}m, Fitzgerald, Nicholson, Hagen, Pasechnik, Olivetti, Martin, Wieser, Silva, Lenders, Wilhelm, Young, Price, Ingold, Allen, Lee, Audren, Probst, Dietrich, Silterra, Webber, Slavi{\v c}, Nothman, Buchner, Kulick, Sch{\"o}nberger, De~Miranda~Cardoso, Reimer, Harrington, Rodr{\'i}guez, {Nunez-Iglesias}, Kuczynski, Tritz, Thoma, Newville, K{\"u}mmerer, Bolingbroke, Tartre, Pak, Smith, Nowaczyk, Shebanov, Pavlyk, Brodtkorb, Lee, McGibbon, Feldbauer, Lewis, Tygier, Sievert, Vigna, Peterson, More, Pudlik, Oshima, Pingel, Robitaille, Spura, Jones, Cera, Leslie, Zito, Krauss, Upadhyay, Halchenko, and {V{\'a}zquez-Baeza}}]{virtanen_scipy_2020}
\bibinfo{author}{P.~Virtanen}, \bibinfo{author}{R.~Gommers}, \bibinfo{author}{T.~E. Oliphant}, \bibinfo{author}{M.~Haberland}, \bibinfo{author}{T.~Reddy}, \bibinfo{author}{D.~Cournapeau}, \bibinfo{author}{E.~Burovski}, \bibinfo{author}{P.~Peterson}, \bibinfo{author}{W.~Weckesser}, \bibinfo{author}{J.~Bright}, \bibinfo{author}{S.~J. Van Der~Walt}, \bibinfo{author}{M.~Brett}, \bibinfo{author}{J.~Wilson}, \bibinfo{author}{K.~J. Millman}, \bibinfo{author}{N.~Mayorov}, \bibinfo{author}{A.~R.~J. Nelson}, \bibinfo{author}{E.~Jones}, \bibinfo{author}{R.~Kern}, \bibinfo{author}{E.~Larson}, \bibinfo{author}{C.~J. Carey}, \bibinfo{author}{{\.I}.~Polat}, \bibinfo{author}{Y.~Feng}, \bibinfo{author}{E.~W. Moore}, \bibinfo{author}{J.~VanderPlas}, \bibinfo{author}{D.~Laxalde}, \bibinfo{author}{J.~Perktold}, \bibinfo{author}{R.~Cimrman}, \bibinfo{author}{I.~Henriksen}, \bibinfo{author}{E.~A. Quintero}, \bibinfo{author}{C.~R. Harris}, \bibinfo{author}{A.~M. Archibald}, \bibinfo{author}{A.~H. Ribeiro}, \bibinfo{author}{F.~Pedregosa}, \bibinfo{author}{P.~Van~Mulbregt}, \bibinfo{author}{{SciPy 1.0 Contributors}}, \bibinfo{author}{A.~Vijaykumar}, \bibinfo{author}{A.~P. Bardelli}, \bibinfo{author}{A.~Rothberg}, \bibinfo{author}{A.~Hilboll}, \bibinfo{author}{A.~Kloeckner}, \bibinfo{author}{A.~Scopatz}, \bibinfo{author}{A.~Lee}, \bibinfo{author}{A.~Rokem}, \bibinfo{author}{C.~N. Woods}, \bibinfo{author}{C.~Fulton}, \bibinfo{author}{C.~Masson}, \bibinfo{author}{C.~H{\"a}ggstr{\"o}m}, \bibinfo{author}{C.~Fitzgerald}, \bibinfo{author}{D.~A. Nicholson}, \bibinfo{author}{D.~R. Hagen}, \bibinfo{author}{D.~V. Pasechnik}, \bibinfo{author}{E.~Olivetti}, \bibinfo{author}{E.~Martin}, \bibinfo{author}{E.~Wieser}, \bibinfo{author}{F.~Silva}, \bibinfo{author}{F.~Lenders}, \bibinfo{author}{F.~Wilhelm}, \bibinfo{author}{G.~Young}, \bibinfo{author}{G.~A. Price}, \bibinfo{author}{G.-L. Ingold}, \bibinfo{author}{G.~E. Allen}, \bibinfo{author}{G.~R. Lee}, \bibinfo{author}{H.~Audren}, \bibinfo{author}{I.~Probst}, \bibinfo{author}{J.~P. Dietrich}, \bibinfo{author}{J.~Silterra}, \bibinfo{author}{J.~T. Webber}, \bibinfo{author}{J.~Slavi{\v c}}, \bibinfo{author}{J.~Nothman}, \bibinfo{author}{J.~Buchner}, \bibinfo{author}{J.~Kulick}, \bibinfo{author}{J.~L. Sch{\"o}nberger}, \bibinfo{author}{J.~V. De~Miranda~Cardoso}, \bibinfo{author}{J.~Reimer}, \bibinfo{author}{J.~Harrington}, \bibinfo{author}{J.~L.~C. Rodr{\'i}guez}, \bibinfo{author}{J.~{Nunez-Iglesias}}, \bibinfo{author}{J.~Kuczynski}, \bibinfo{author}{K.~Tritz}, \bibinfo{author}{M.~Thoma}, \bibinfo{author}{M.~Newville}, \bibinfo{author}{M.~K{\"u}mmerer}, \bibinfo{author}{M.~Bolingbroke}, \bibinfo{author}{M.~Tartre}, \bibinfo{author}{M.~Pak}, \bibinfo{author}{N.~J. Smith}, \bibinfo{author}{N.~Nowaczyk}, \bibinfo{author}{N.~Shebanov}, \bibinfo{author}{O.~Pavlyk}, \bibinfo{author}{P.~A. Brodtkorb}, \bibinfo{author}{P.~Lee}, \bibinfo{author}{R.~T. McGibbon}, \bibinfo{author}{R.~Feldbauer}, \bibinfo{author}{S.~Lewis}, \bibinfo{author}{S.~Tygier}, \bibinfo{author}{S.~Sievert}, \bibinfo{author}{S.~Vigna}, \bibinfo{author}{S.~Peterson}, \bibinfo{author}{S.~More}, \bibinfo{author}{T.~Pudlik}, \bibinfo{author}{T.~Oshima}, \bibinfo{author}{T.~J. Pingel}, \bibinfo{author}{T.~P. Robitaille}, \bibinfo{author}{T.~Spura}, \bibinfo{author}{T.~R. Jones}, \bibinfo{author}{T.~Cera}, \bibinfo{author}{T.~Leslie}, \bibinfo{author}{T.~Zito}, \bibinfo{author}{T.~Krauss}, \bibinfo{author}{U.~Upadhyay}, \bibinfo{author}{Y.~O. Halchenko}, \bibinfo{author}{Y.~{V{\'a}zquez-Baeza}},
\newblock \bibinfo{title}{{{SciPy}} 1.0: Fundamental algorithms for scientific computing in {{Python}}},
\newblock \bibinfo{journal}{Nature Methods} \bibinfo{volume}{17} (\bibinfo{year}{2020}) \bibinfo{pages}{261--272}. \DOIprefix\doi{10.1038/s41592-019-0686-2}.
\bibitem[{Box and Pierce(1970)}]{box_distribution_1970}
\bibinfo{author}{G.~E.~P. Box}, \bibinfo{author}{D.~A. Pierce},
\newblock \bibinfo{title}{Distribution of {{Residual Autocorrelations}} in {{Autoregressive-Integrated Moving Average Time Series Models}}},
\newblock \bibinfo{journal}{Journal of the American Statistical Association} \bibinfo{volume}{65} (\bibinfo{year}{1970}) \bibinfo{pages}{1509--1526}. \DOIprefix\doi{10.2307/2284333}. \href{http://arxiv.org/abs/2284333}{{\tt arXiv:2284333}}.
\bibitem[{Hyndman et~al.(2008)Hyndman, Koehler, Ord, and Snyder}]{hyndman_forecasting_2008}
\bibinfo{author}{R.~Hyndman}, \bibinfo{author}{A.~B. Koehler}, \bibinfo{author}{J.~K. Ord}, \bibinfo{author}{R.~D. Snyder}, \bibinfo{title}{Forecasting with {{Exponential Smoothing}}: {{The State Space Approach}}}, \bibinfo{publisher}{{Springer Science \& Business Media}}, \bibinfo{year}{2008}.
\bibitem[{Assimakopoulos and Nikolopoulos(2000)}]{assimakopoulos_theta_2000}
\bibinfo{author}{V.~Assimakopoulos}, \bibinfo{author}{K.~Nikolopoulos},
\newblock \bibinfo{title}{The theta model: A decomposition approach to forecasting},
\newblock \bibinfo{journal}{International Journal of Forecasting} \bibinfo{volume}{16} (\bibinfo{year}{2000}) \bibinfo{pages}{521--530}. \DOIprefix\doi{10.1016/S0169-2070(00)00066-2}.
\bibitem[{Croston(1972)}]{croston_forecasting_1972}
\bibinfo{author}{J.~D. Croston},
\newblock \bibinfo{title}{Forecasting and {{Stock Control}} for {{Intermittent Demands}}},
\newblock \bibinfo{journal}{Operational Research Quarterly (1970-1977)} \bibinfo{volume}{23} (\bibinfo{year}{1972}) \bibinfo{pages}{289--303}. \DOIprefix\doi{10.2307/3007885}. \href{http://arxiv.org/abs/3007885}{{\tt arXiv:3007885}}.
\bibitem[{Kunz et~al.(2023)Kunz, Birr, Raslan, Ma, Li, Gouttes, Koren, Naghibi, Stephan, Bulycheva, Grzeschik, Keki{\'c}, Narodovitch, Rasul, Sieber, and Januschowski}]{kunz_deep_2023}
\bibinfo{author}{M.~Kunz}, \bibinfo{author}{S.~Birr}, \bibinfo{author}{M.~Raslan}, \bibinfo{author}{L.~Ma}, \bibinfo{author}{Z.~Li}, \bibinfo{author}{A.~Gouttes}, \bibinfo{author}{M.~Koren}, \bibinfo{author}{T.~Naghibi}, \bibinfo{author}{J.~Stephan}, \bibinfo{author}{M.~Bulycheva}, \bibinfo{author}{M.~Grzeschik}, \bibinfo{author}{A.~Keki{\'c}}, \bibinfo{author}{M.~Narodovitch}, \bibinfo{author}{K.~Rasul}, \bibinfo{author}{J.~Sieber}, \bibinfo{author}{T.~Januschowski}, \bibinfo{title}{Deep {{Learning}} based {{Forecasting}}: A case study from the online fashion industry}, \bibinfo{year}{2023}. \DOIprefix\doi{10.48550/arXiv.2305.14406}. \href{http://arxiv.org/abs/2305.14406}{{\tt arXiv:2305.14406}}.
\bibitem[{Lim et~al.(2021)Lim, Ar{\i}k, Loeff, and Pfister}]{lim_temporal_2021}
\bibinfo{author}{B.~Lim}, \bibinfo{author}{S.~{\"O}. Ar{\i}k}, \bibinfo{author}{N.~Loeff}, \bibinfo{author}{T.~Pfister},
\newblock \bibinfo{title}{Temporal {{Fusion Transformers}} for interpretable multi-horizon time series forecasting},
\newblock \bibinfo{journal}{International Journal of Forecasting} \bibinfo{volume}{37} (\bibinfo{year}{2021}) \bibinfo{pages}{1748--1764}. \DOIprefix\doi{10.1016/j.ijforecast.2021.03.012}.
\bibitem[{Li et~al.(2019)Li, Jin, Xuan, Zhou, Chen, Wang, and Yan}]{li_enhancing_2019}
\bibinfo{author}{S.~Li}, \bibinfo{author}{X.~Jin}, \bibinfo{author}{Y.~Xuan}, \bibinfo{author}{X.~Zhou}, \bibinfo{author}{W.~Chen}, \bibinfo{author}{Y.-X. Wang}, \bibinfo{author}{X.~Yan},
\newblock \bibinfo{title}{Enhancing the {{Locality}} and {{Breaking}} the {{Memory Bottleneck}} of {{Transformer}} on {{Time Series Forecasting}}},
\newblock in: \bibinfo{booktitle}{Advances in {{Neural Information Processing Systems}} 32}, \bibinfo{publisher}{{Curran Associates, Inc.}}, \bibinfo{year}{2019}, pp. \bibinfo{pages}{5244--5254}.
\bibitem[{Akiba et~al.(2019)Akiba, Sano, Yanase, Ohta, and Koyama}]{akiba_optuna_2019}
\bibinfo{author}{T.~Akiba}, \bibinfo{author}{S.~Sano}, \bibinfo{author}{T.~Yanase}, \bibinfo{author}{T.~Ohta}, \bibinfo{author}{M.~Koyama},
\newblock \bibinfo{title}{Optuna: {{A Next-generation Hyperparameter Optimization Framework}}},
\newblock in: \bibinfo{booktitle}{Proceedings of the 25th {{ACM SIGKDD International Conference}} on {{Knowledge Discovery}} \& {{Data Mining}}}, {{KDD}} '19, \bibinfo{publisher}{{Association for Computing Machinery}}, \bibinfo{address}{{New York, NY, USA}}, \bibinfo{year}{2019}, pp. \bibinfo{pages}{2623--2631}. \DOIprefix\doi{10.1145/3292500.3330701}.
\bibitem[{Garza et~al.(2022)Garza, Mergenthaler~Canseco, Chall{\'u}, and Olivares}]{garza_statsforecast_2022}
\bibinfo{author}{F.~Garza}, \bibinfo{author}{M.~Mergenthaler~Canseco}, \bibinfo{author}{C.~Chall{\'u}}, \bibinfo{author}{K.~G. Olivares},
\newblock \bibinfo{title}{{{StatsForecast}}: {{Lightning}} fast forecasting with statistical and econometric models.},
\newblock in: \bibinfo{booktitle}{{{PyCon}}}, \bibinfo{address}{{Salt Lake City, USA}}, \bibinfo{year}{2022}.
\bibitem[{Sprangers et~al.(2021)Sprangers, Schelter, and {de Rijke}}]{sprangers_probabilistic_2021}
\bibinfo{author}{O.~Sprangers}, \bibinfo{author}{S.~Schelter}, \bibinfo{author}{M.~{de Rijke}},
\newblock \bibinfo{title}{Probabilistic {{Gradient Boosting Machines}} for {{Large-Scale Probabilistic Regression}}},
\newblock in: \bibinfo{booktitle}{Proceedings of the 27th {{ACM SIGKDD Conference}} on {{Knowledge Discovery}} \& {{Data Mining}}}, {{KDD}} '21, \bibinfo{publisher}{{Association for Computing Machinery}}, \bibinfo{address}{{New York, NY, USA}}, \bibinfo{year}{2021}, pp. \bibinfo{pages}{1510--1520}. \DOIprefix\doi{10.1145/3447548.3467278}.
\bibitem[{Hasson et~al.(2021)Hasson, Wang, Januschowski, and Gasthaus}]{hasson_probabilistic_2021}
\bibinfo{author}{H.~Hasson}, \bibinfo{author}{B.~Wang}, \bibinfo{author}{T.~Januschowski}, \bibinfo{author}{J.~Gasthaus},
\newblock \bibinfo{title}{Probabilistic {{Forecasting}}: {{A Level-Set Approach}}},
\newblock in: \bibinfo{booktitle}{Advances in {{Neural Information Processing Systems}}}, volume~\bibinfo{volume}{34}, \bibinfo{publisher}{{Curran Associates, Inc.}}, \bibinfo{year}{2021}, pp. \bibinfo{pages}{6404--6416}.
\bibitem[{Stankeviciute et~al.(2021)Stankeviciute, M.~Alaa, and {van der Schaar}}]{stankeviciute_conformal_2021}
\bibinfo{author}{K.~Stankeviciute}, \bibinfo{author}{A.~M.~Alaa}, \bibinfo{author}{M.~{van der Schaar}},
\newblock \bibinfo{title}{Conformal {{Time-series Forecasting}}},
\newblock in: \bibinfo{booktitle}{Advances in {{Neural Information Processing Systems}}}, volume~\bibinfo{volume}{34}, \bibinfo{publisher}{{Curran Associates, Inc.}}, \bibinfo{year}{2021}, pp. \bibinfo{pages}{6216--6228}.

\end{thebibliography}

\clearpage

\appendix
\onecolumn

\rev{\section{Derivation of gradient and second-order derivative} \label{app:hl_derivation}
We have the following loss function (Eq.~\eqref{eq:whloss}):  
\begin{align} 
  L &= \sum \left[ \frac{1}{2} \left( \left(\textbf{Y} - \tilde{\textbf{Y}} \right) \odot \left(\textbf{Y} - \tilde{\textbf{Y}} \right) \right) \oslash \left(d^{cs} d^{te}\right) \right] \; .
\end{align}
First, note that \(L\) is a scalar, as we sum over all \(n^{cs}\; \times \; n^{te}\) elements of the matrix contained in the summation. We can expand this summation as follows:
\begin{align}
  L &= \frac{\left( \frac{1}{2} \left(\textbf{Y}_{0, 0} - \tilde{\textbf{Y}}_{0, 0} \right)^2 \right)}{\left(d^{cs} d^{te}\right)_{0, 0}} + \frac{\left( \frac{1}{2} \left(\textbf{Y}_{1, 0} - \tilde{\textbf{Y}}_{1, 0} \right)^2 \right)}{\left(d^{cs} d^{te}\right)_{1, 0}} + \dots + \frac{\left( \frac{1}{2} \left(\textbf{Y}_{n^{cs}, n^{te}} - \tilde{\textbf{Y}}_{n^{cs}, n^{te}} \right)^2 \right)}{\left(d^{cs} d^{te}\right)_{n^{cs}, n^{te}}} \; ,
\end{align}
with \(\textbf{Y}_{i, j}\), \(\tilde{\textbf{Y}}_{i, j}\) and \(\left(d^{cs} d^{te}\right)_{i, j}\) denoting the \(i, j\)-th entry of the matrices \(\textbf{Y}\), \(\tilde{\textbf{Y}}\) and \(\left(d^{cs} d^{te}\right)\), respectively. The derivative of \(L\) with respect to the first element \(\tilde{\textbf{Y}}_{0, 0}\) is the standard squared error loss gradient divided by \(\left(d^{cs} d^{te}\right)_{1,1}\), as all other terms in the summation cancel out:
\begin{align}
  \frac{\partial L}{\partial \tilde{\textbf{Y}}_{0, 0}} &= \frac{\left(\tilde{\textbf{Y}}_{0, 0}  - \textbf{Y}_{0, 0}  \right)}{\left(d^{cs} d^{te}\right)_{0, 0}} \; .
\end{align}
The second derivative of \(L\) with respect to the first element \(\tilde{\textbf{Y}}_{0, 0}\) then reads:
\begin{align}
  \frac{\partial^2 L}{\partial \left(\tilde{\textbf{Y}}_{0, 0}\right)^2} &= \frac{1}{\left(d^{cs} d^{te}\right)_{0, 0}} \; .
\end{align}
We can then straightforwardly generalize the first and second derivative to the matrix case, such that the first and second derivative of \(L\) with respect to the matrix \( \tilde{\textbf{Y}}\) read:
\begin{align}
  \frac{\partial L}{\partial \tilde{\textbf{Y}}{}} &= \left(\left(\tilde{\textbf{Y}} - \textbf{Y} \right) \oslash \left(d^{cs} d^{te}\right) \right) \;, \\
  \frac{\partial^2 L}{\partial \left(\tilde{\textbf{Y}}{}\right)^2} &= \left(\textbf{1} \oslash \left(d^{cs} d^{te}\right) \right) \;,
\end{align}
with \(\textbf{1}\) denoting a matrix of ones of the same size as \(\tilde{\textbf{Y}}\). As \(L\) is a scalar and \(\tilde{\textbf{Y}} \in \mathbb{R}^{n^{cs} \; \times \; n^{te}}\), the derivative \(\frac{\partial L}{\partial \tilde{\textbf{Y}}{}}\) is a matrix of the same size as \(\tilde{\textbf{Y}}\). Similarly, \(\frac{\partial^2 L}{\partial \left(\tilde{\textbf{Y}}{}\right)^2}\) is a matrix of the same size as \(\tilde{\textbf{Y}}\), as we only consider the second-order derivatives with respect to the same component in \(\tilde{\textbf{Y}}\). To get to the gradient of \(L\) with respect to the bottom-level forecasts \(\hat{\textbf{Y}}{}^{n_b}\), we need to collect every gradient component in \(\frac{\partial L}{\partial \tilde{\textbf{Y}}{}}\) that contains a bottom-level forecast element. We can achieve this by multiplying the derivatives by our summing matrices \(S^{cs}\) and \(S^{te}\):
\begin{align}
  \underbrace{\frac{\partial L}{\partial \hat{\textbf{Y}}{}^{n_b}}}_{\mathbb{R}^{n_b^{cs} \; \times \; n_b^{te}}} &= \underbrace{(S^{cs})^\intercal}_{\mathbb{R}^{n_b^{cs} \; \times \; n^{cs}}} \underbrace{\left(\frac{\partial L}{\partial \tilde{\textbf{Y}}}\right)}_{\mathbb{R}^{n^{cs} \; \times \; n^{te}}} \underbrace{S^{te}}_{\mathbb{R}^{n^{te} \; \times \; n_b^{te}}} \;, \\
  \underbrace{\frac{\partial^2 L}{\partial \left(\hat{\textbf{Y}}{}^{n_b} \right)^2}}_{\mathbb{R}^{n_b^{cs} \; \times \; n_b^{te}}} &= \underbrace{(S^{cs})^\intercal}_{\mathbb{R}^{n_b^{cs} \; \times \; n^{cs}}} \underbrace{\left( \frac{\partial^2 L}{\partial \left(\tilde{\textbf{Y}}{}\right)^2} \right)}_{\mathbb{R}^{n^{cs} \; \times \; n^{te}}} \underbrace{S^{te}}_{\mathbb{R}^{n^{te} \; \times \; n_b^{te}}}  \;.
\end{align}
}

\section{Derivation of gradient of \rev{toy example}} \label{app:hl_derivation_toy}
We have the following hierarchical forecasting problem:
\rev{
\begin{align*}
  \tilde{\textbf{Y}} = \underbrace{
    \begin{bmatrix}
    1 &1 \\
    1 &0 \\
    0 &1
    \end{bmatrix}}_{S^{cs}}
    \underbrace{    
    \begin{bmatrix}
      \hat{\textbf{y}}_{0, 0} & \hat{\textbf{y}}_{0, 1} \\
      \hat{\textbf{y}}_{1, 0} & \hat{\textbf{y}}_{1, 1} \\
    \end{bmatrix}}_{\hat{\textbf{Y}}{}^{n_b}}
    \underbrace{
    \begin{bmatrix}
      1 &1 &0\\
      1 &0 &1\\
      \end{bmatrix}}_{(S^{te})^\intercal}     
    \;,
\end{align*}}
which results in the following gradient:
\rev{
\begin{align*} 
  \begin{bmatrix}
  \frac{\partial L}{\partial \hat{\textbf{y}}_{0, 0}} & \frac{\partial L}{\partial \hat{\textbf{y}}_{0, 1}} \\
  \frac{\partial L}{\partial \hat{\textbf{y}}_{1, 0}} & \frac{\partial L}{\partial \hat{\textbf{y}}_{1, 1}}
  \end{bmatrix} 
  &= (S^{cs})^\intercal \left[ \left(S^{cs} \left(\hat{\textbf{Y}}{}^{n_b} - \textbf{Y}{}^{n_b}\right) (S^{te})^\intercal\right) \oslash \left( d^{cs} d^{te} \right) \right] S^{te} \\
  &= (S^{cs})^\intercal \left[ \left(S^{cs} \left(\hat{\textbf{Y}}{}^{n_b} - \textbf{Y}{}^{n_b}\right) (S^{te})^\intercal\right) \oslash \left( \left(l^{cs} S^{cs} \textbf{1}^{cs}\right) \left(l^{te} S^{te} \textbf{1}^{te}\right)^\intercal \right) \right] S^{te} \\
  &= \begin{bmatrix}
    1 &1 &0\\
    1 &0 &1\\
    \end{bmatrix}     
    \left[ \left(
    \begin{bmatrix}
      1 &1 \\
      1 &0 \\
      0 &1
    \end{bmatrix}
    \left(  
    \begin{bmatrix}
      \hat{\textbf{y}}_{0, 0} &\hat{\textbf{y}}_{0, 1} \\
      \hat{\textbf{y}}_{1, 0} &\hat{\textbf{y}}_{1, 1}
    \end{bmatrix}  
    - 
    \begin{bmatrix}
      \textbf{y}_{0, 0} &\textbf{y}_{0, 1} \\
      \textbf{y}_{1, 0} &\textbf{y}_{1, 1}
    \end{bmatrix}  
    \right)
    \begin{bmatrix}
      1 &1 &0\\
      1 &0 &1\\
    \end{bmatrix} 
    \right) \oslash \left(\left( 2 
    \begin{bmatrix}
      2 &1 &1
    \end{bmatrix}^\intercal 
    \right) 
    \left( 2 
    \begin{bmatrix}
      2 &1 &1
    \end{bmatrix}
    \right)
    \right) \right]
    \begin{bmatrix}
      1 &1 \\
      1 &0 \\
      0 &1
    \end{bmatrix}  \;, \nonumber \\
    &= \begin{bmatrix}
      1 &1 &0\\
      1 &0 &1\\
      \end{bmatrix}     
      \left[ \left(
      \begin{bmatrix}
        1 &1 \\
        1 &0 \\
        0 &1
      \end{bmatrix}
        \begin{bmatrix}
          \overbrace{\hat{\textbf{y}}_{0, 0} - \textbf{y}_{0, 0}}^{e_{0, 0}} &\overbrace{\hat{\textbf{y}}_{0, 1} - \textbf{y}_{0, 1}}^{e_{0, 1}} \\
          \underbrace{\hat{\textbf{y}}_{1, 0} - \textbf{y}_{1, 0}}_{e_{1, 0}} &\underbrace{\hat{\textbf{y}}_{1, 1} - \textbf{y}_{1, 1}}_{e_{1, 1}}
        \end{bmatrix}  
      \begin{bmatrix}
        1 &1 &0\\
        1 &0 &1\\
      \end{bmatrix} 
      \right) \oslash 
      \begin{bmatrix}
        16 &8 &8 \\
        8 &4 &4 \\
        8 &4 &4 \\
      \end{bmatrix}
      \right]
      \begin{bmatrix}
        1 &1 \\
        1 &0 \\
        0 &1
      \end{bmatrix}  \;, \nonumber \\
      &= \begin{bmatrix}
        1 &1 &0\\
        1 &0 &1\\
        \end{bmatrix}     
        \begin{bmatrix}
          \frac{1}{16} (e_{0, 0} + e_{1, 0} + e_{0, 1} + e_{1, 1}) &\frac{1}{8} (e_{0, 0} + e_{1, 0}) &\frac{1}{8} (e_{0, 1} + e_{1, 1}) \\
          \frac{1}{8} (e_{0, 0} + e_{0, 1})  &\frac{1}{4} e_{0, 0} & \frac{1}{4} e_{0, 1} \\
          \frac{1}{8} (e_{1, 0} + e_{1, 1})  &\frac{1}{4} e_{1, 0} & \frac{1}{4} e_{1, 1} \\
        \end{bmatrix}
        \begin{bmatrix}
          1 &1 \\
          1 &0 \\
          0 &1
        \end{bmatrix}  \;, \nonumber \\      
    \begin{split}
      &= 
      \left[
      \begin{matrix}
        \frac{1}{16} (e_{0, 0} + e_{1, 0} + e_{0, 1} + e_{1, 1}) + \frac{1}{8} (e_{0, 0} + e_{1, 0}) + \frac{1}{8} (e_{0, 0} + e_{0, 1}) + \frac{1}{4} e_{0, 0}  \\
        \frac{1}{16} (e_{0, 0} + e_{1, 0} + e_{0, 1} + e_{1, 1}) + \frac{1}{8} (e_{0, 0} + e_{1, 0}) + \frac{1}{8} (e_{1, 0} + e_{1, 1}) + \frac{1}{4} e_{1, 0} 
      \end{matrix} \right. \\
      &\quad
      \left.
      \begin{matrix}
        \frac{1}{16} (e_{0, 0} + e_{1, 0} + e_{0, 1} + e_{1, 1}) + \frac{1}{8} (e_{0, 1} + e_{1, 1}) + \frac{1}{8} (e_{0, 0} + e_{0, 1}) + \frac{1}{4} e_{0, 1} \\
        \frac{1}{16} (e_{0, 0} + e_{1, 0} + e_{0, 1} + e_{1, 1}) + \frac{1}{8} (e_{0, 1} + e_{1, 1}) + \frac{1}{8} (e_{1, 0} + e_{1, 1}) + \frac{1}{4} e_{1, 1}
      \end{matrix} \right] \;, \\      
     \end{split}
      \nonumber \\   
      &= 
      \begin{bmatrix}
        \frac{9}{16}e_{0, 0} + \frac{3}{16}e_{1, 0} + \frac{3}{16}e_{0, 1} + \frac{1}{16}e_{1, 1} & \frac{9}{16}e_{0, 1} + \frac{3}{16}e_{0, 0} + \frac{3}{16}e_{1,1} + \frac{1}{16}e_{1, 0} \\
        \frac{9}{16}e_{1, 0} + \frac{3}{16}e_{0, 0} + \frac{3}{16}e_{1, 1} + \frac{1}{16}e_{0, 1} & \frac{9}{16}e_{1, 1} + \frac{3}{16}e_{1, 0} + \frac{3}{16}e_{0, 1} + \frac{1}{16}e_{0, 0} \\
      \end{bmatrix}  \;. \nonumber \\     
 \end{align*}}  

\section{M5 Dataset} \label{app:m5dataset}
For each of the scenarios of our experiments in Section~\ref{sec:experiments}, we construct a set of features for the LightGBM model as given in Table~\ref{tab:features}. To facilitate the most `fair' comparison across methods, each model has the same features, and for the time series aggregations in the hierarchy we construct the features taken over the aggregation. \rev{We refer the reader to \cite{makridakis_m5_2021} for a detailed description of the dataset.}
\begin{table}
  \caption{Features used for the M5 dataset in our experiments.}
  \label{tab:features}
  \centering
  {\small\setlength{\tabcolsep}{1pt} 
  \begin{tabular}{l l }
  \toprule 
  Feature & Description \\
  \midrule
  \texttt{Aggregation} & Aggregation level in the hierarchy \\
  \texttt{Value} & Identifier of time series of this aggregation \\   
  \texttt{sales\_lag1-7} & Lagged sales (target) (7 features) \\
  \texttt{sales\_lag28} & Sales 28 days ago \\ 
  \texttt{sales\_lag56} & Sales 56 days ago \\ 
  \texttt{sales\_lag364} & Sales last year \\ 
  \texttt{sales\_lag1\_mavg7} & Moving average of sales last 7 days \\
  \texttt{sales\_lag1\_mavg28} & Moving average of sales last 28 days \\ 
  \texttt{sales\_lag1\_mavg56} & Moving average of sales last 56 days\\ 
  \texttt{dayofweek} & Day of the week \\ 
  \texttt{dayofmonth} & Day of the month \\
  \texttt{weekofyear} & Week of year \\ 
  \texttt{monthofyear} & Month of year \\ 
  \texttt{sell\_price\_avg} & Sell price (average if aggregation)\\ 
  \texttt{sell\_price\_change} & Day-to-day change in sell price\\
  \texttt{weeks\_on\_sale\_avg} & Weeks on sale \\ 
  \texttt{snap\_CA} & State indicator for California \\ 
  \texttt{snap\_TX} & State indicator for Texas \\ 
  \texttt{snap\_WI} & State indicator for Wyoming \\
  \texttt{event\_type\_1\_enc} & Encoded events \\ 
  \texttt{event\_type\_2\_enc} & Encoded events \\
  \bottomrule
  \end{tabular}}
\end{table}

\section{M5 model training \& optimization} \label{app:hyperparam}
We optimize the hyperparameters of our LightGBM models using Optuna \cite{akiba_optuna_2019}, using the settings found in Table~\ref{tab:hyperparams}. The validation is performed on a rolling-forward basis for 3 validation sets, where we use three years of data to predict the next 28 days ahead. After the hyperparameter optimization procedure, we use the average number of iterations at which the lowest validation loss was achieved across the 3 validation sets as the number of estimators to use in our final model. The final model uses the last three years of data preceding the first day in the test set.
\begin{table*}
  \caption{Key hyperparameters used in our experiments. The parameters with a search range included are optimized in a hyperparameter search.}
  \label{tab:hyperparams}
  \centering
  {\small\setlength{\tabcolsep}{1pt} 
  \begin{tabular}{l l c l }
  \toprule 
  Parameter & Description & Default value & Search range \\
  \midrule
  \texttt{n\_estimators} & Number of trees in each model & 2000 & Lowest validation loss \\
  \texttt{n\_trials} & Number of optimization trials to run & 100 & \\  
  \texttt{learning\_rate} & Learning rate & 0.05 & \\  
  \texttt{n\_validation\_sets} & Number of validation sets & 3 & \\  
  \texttt{n\_days\_test} & Number of days in validation and test sets & 28 & \\
  \texttt{max\_levels\_random} & Max. number of levels when using a random hierarchy & \revv{10} & \\  
  \texttt{\shortstack{max\_categories\_per \\ \_random\_level}} & Max. categories per level in the random hierarchy & \revv{100} & \\  
  \texttt{hier\_freq} & Frequency of performing the randomized hierarchical aggregation & 1 & \texttt{uniform($1$, $10$)}  \\  
  \texttt{lambda\_l1} & L1-regularization & 0 & \texttt{log\_uniform($10^{-8}$, $10^{1}$)} \\  
  \texttt{lambda\_l2} & L2-regularization & 0 & \texttt{log\_uniform($10^{-8}$, $10^{1}$)} \\  
  \texttt{num\_leaves} & Max. number of leaves per tree & 31 & \texttt{uniform($2^{3}$, $2^{10}$)} \\  
  \texttt{feature\_fraction} & Fraction of features to use to build a tree & 1.0 & \texttt{uniform($0.4$, $1.0$)} \\  
  \texttt{bagging\_fraction} & Fraction of training samples to use to build a tree & 1.0 & \texttt{uniform($0.4$, $1.0$)} \\  
  \texttt{bagging\_freq} & Frequency at which to create a new bagging batch & 1.0 & \texttt{uniform($1$, $7$)} \\  
  \texttt{min\_child\_samples} & Minimum number of samples per leaf & 20 & \texttt{log\_uniform($5$, $5000$)} \\  
  \bottomrule
  \end{tabular}}
\end{table*}

\section{Experiments} \label{app:experiments}

\begin{sidewaystable*}[t]
  \caption{Forecasting results for all stores on the M5 dataset. We report mean RMSE scores across 10 random seeds with standard deviation in brackets. Lower is better.}
  \label{tab:allstores_abs_rmse}
  \centering\scalebox{0.85}{
  {\small\setlength{\tabcolsep}{1pt} 
  \begin{tabular}{l l l rrrrrrrrrrrrr}
  \toprule 
   &&&& &  &\multicolumn{3}{ c }{Store}   &\multicolumn{2}{ c }{Product} &\multicolumn{3}{ c }{State} \\
   \cmidrule(r){7-9} \cmidrule(r){10-11} \cmidrule(r){12-14}
  Scen./Obj. & Metric  & Reconciliation &Product	&Department	&Category &Department	&Category	&Total &Store	&State &Department &Category &Total	&Total	&All series \\
  \midrule																	
  \texttt{Bottom-up}																	\\
  \hspace{0.1cm} 	SL	&SL	&None	&2.11 (0.00)	&580 (3.95)	&1096 (13.71)	&110 (0.60)	&207 (1.21)	&416 (2.07)	&8.8 (0.03)	&4.26 (0.01)	&263 (1.30)	&510 (3.81)	&997 (8.44)	&2086 (38.53)	&22.4 (0.16)	\\
  \hspace{0.1cm} 	SL	&HL	&None	&2.11 (0.00)	&564 (5.58)	&1039 (14.64)	&107 (0.74)	&200 (1.69)	&408 (2.37)	&8.76 (0.06)	&4.25 (0.01)	&256 (1.30)	&490 (3.93)	&983 (5.93)	&2079 (33.46)	&21.8 (0.15)	\\
  \hspace{0.1cm} 	HL	&HL	&None	&2.11 (0.00)	&511 (8.93)	&882 (20.54)	&102 (0.90)	&185 (1.48)	&388 (4.21)	&8.5 (0.02)	&4.21 (0.01)	&233 (2.42)	&427 (4.78)	&878 (17.17)	&1819 (78.42)	&19.5 (0.38)	\\
  \hspace{0.1cm} 	HL	&SL	&None	&2.11 (0.00)	&511 (9.47)	&886 (24.67)	&103 (1.08)	&187 (2.12)	&390 (3.80)	&8.46 (0.04)	&4.19 (0.01)	&234 (3.15)	&430 (7.18)	&881 (15.27)	&1813 (64.6)	&19.6 (0.36)	\\
  \hspace{0.1cm} 	TL	&HL	&None	&2.11 (0.00)	&549 (9.21)	&1055 (22.68)	&109 (0.58)	&207 (1.27)	&415 (2.58)	&8.73 (0.03)	&4.25 (0.01)	&256 (2.09)	&502 (5.04)	&978 (13.8)	&2007 (63.57)	&21.8 (0.34)	\\
  \hspace{0.1cm} 	TL	&SL	&None	&2.11 (0.00)	&544 (7.66)	&1017 (25.23)	&110 (0.40)	&208 (1.20)	&416 (2.41)	&8.68 (0.04)	&4.25 (0.01)	&256 (1.55)	&497 (5.23)	&972 (7.67)	&1943 (41.21)	&21.5 (0.25)	\\
  \hspace{0.1cm} 	TL	&TL	&None	&2.46 (0.00)	&1579 (9.71)	&3075 (20.5)	&193 (0.87)	&379 (1.84)	&717 (5.17)	&13.41 (0.05)	&5.68 (0.02)	&566 (3.11)	&1111 (6.42)	&2074 (18.68)	&5646 (60.40)	&53.3 (0.39)	\\
  \midrule																	
  \texttt{Sep. agg.}																	\\
  \hspace{0.1cm} 	SL	&SL	&Base	&2.11 (0.00)	&835 (23.47)	&1417 (69.78)	&130 (1.86)	&236 (5.25)	&474 (8.71)	&8.88 (0.02)	&4.23 (0.01)	&322 (11.48)	&682 (19.02)	&1269 (47.11)	&3339 (86.79)	&30.1 (0.21)	\\
  \hspace{0.1cm} 	SL	&SL	&OLS	&2.11 (0.00)	&804 (14.22)	&1548 (22.24)	&120 (1.49)	&220 (1.95)	&443 (2.37)	&8.77 (0.02)	&4.25 (0.01)	&315 (4.15)	&605 (7.98)	&1222 (9.69)	&3125 (24.25)	&29.1 (0.22)	\\
  \hspace{0.1cm} 	SL	&SL	&WLS-struct	&2.10 (0.00)	&731 (8.55)	&1505 (16.09)	&113 (0.58)	&217 (1.13)	&426 (2.17)	&8.71 (0.02)	&4.23 (0.01)	&291 (2.38)	&591 (4.82)	&1153 (7.53)	&2909 (23.52)	&27.6 (0.22)	\\
  \hspace{0.1cm} 	SL	&SL	&WLS-var	&2.11 (0.00)	&647 (7.23)	&1345 (18.95)	&109 (0.51)	&211 (1.43)	&411 (2.06)	&8.72 (0.03)	&4.24 (0.01)	&272 (2.06)	&555 (5.58)	&1068 (8.77)	&2541 (31.06)	&25.1 (0.25)	\\
  \hspace{0.1cm} 	SL	&SL	&MinT-shrink	&2.11 (0.00)	&667 (18.15)	&1397 (44.32)	&106 (1.15)	&206 (3.14)	&405 (5.45)	&8.76 (0.02)	&4.24 (0.01)	&272 (4.92)	&558 (12.59)	&1091 (22.52)	&2706 (80.28)	&25.8 (0.63)	\\
  \hspace{0.1cm} 	SL	&SL	&ERM	&2.58 (0.01)	&728 (48.01)	&1411 (118.36)	&117 (3.66)	&213 (8.73)	&443 (20.5)	&10.30 (0.10)	&5.21 (0.04)	&307 (14.61)	&583 (33.15)	&1215 (75.3)	&3101 (258.29)	&28.1 (1.83)	\\
  \midrule																	
  \texttt{Global}																	\\
  \hspace{0.1cm} 	SL	&SL	&Base	&2.16 (0.00)	&771 (54.94)	&1587 (256.06)	&120 (2.11)	&227 (8.81)	&458 (18.45)	&9.09 (0.05)	&4.4 (0.01)	&329 (18.51)	&650 (31.09)	&1807 (1203.62)	&3273 (338.71)	&32.8 (6.60)	\\
  \hspace{0.1cm} 	SL	&SL	&OLS	&2.13 (0.01)	&764 (58.91)	&1524 (181.38)	&117 (10.32)	&227 (27.24)	&484 (109.14)	&8.99 (0.05)	&4.34 (0.03)	&316 (42.99)	&637 (107.63)	&1371 (417.84)	&3114 (417.38)	&29.9 (2.52)	\\
  \hspace{0.1cm} 	SL	&SL	&WLS-struct	&2.13 (0.00)	&800 (45.37)	&1686 (103.86)	&118 (2.52)	&235 (5.96)	&463 (11.90)	&9.09 (0.03)	&4.35 (0.01)	&314 (10.72)	&651 (24.97)	&1264 (55.94)	&3231 (279.96)	&30.4 (1.68)	\\
  \hspace{0.1cm} 	SL	&SL	&WLS-var	&2.14 (0.01)	&876 (43.62)	&1867 (106.77)	&130 (2.31)	&263 (7.10)	&508 (18.51)	&9.10 (0.05)	&4.35 (0.01)	&345 (11.71)	&727 (30.30)	&1374 (80.66)	&3463 (283.05)	&33.2 (1.84)	\\
  \hspace{0.1cm} 	SL	&SL	&MinT-shrink	&2.18 (0.02)	&730 (50.73)	&1545 (155.65)	&115 (2.42)	&228 (5.97)	&478 (19.61)	&9.29 (0.16)	&4.46 (0.06)	&293 (12.06)	&599 (33.27)	&1235 (89.41)	&3211 (432.02)	&29.1 (2.33)	\\
  \hspace{0.1cm} 	SL	&SL	&ERM	&2.56 (0.14)	&923 (113.39)	&1857 (271.23)	&138 (16.52)	&265 (34.54)	&558 (105.87)	&10.52 (0.62)	&5.23 (0.34)	&382 (56.45)	&762 (118.54)	&1607 (371.34)	&3752 (669.36)	&35.7 (3.74)	\\
  
  \bottomrule
  
\end{tabular}}}
  \end{sidewaystable*}

  \begin{sidewaystable*}[t]
    \caption{Forecasting results for all stores on the M5 dataset. We report mean MAE scores across 10 random seeds with standard deviation in brackets. Lower is better.}
    \label{tab:allstores_abs_mae}
    \centering\scalebox{0.85}{
    {\small\setlength{\tabcolsep}{1pt} 
    \begin{tabular}{l l l rrrrrrrrrrrrr}
    \toprule 
     &&&& &  &\multicolumn{3}{ c }{Store}   &\multicolumn{2}{ c }{Product} &\multicolumn{3}{ c }{State} \\
     \cmidrule(r){7-9} \cmidrule(r){10-11} \cmidrule(r){12-14}
    Scen./Obj. & Metric  & Reconciliation &Product	&Department	&Category &Department	&Category	&Total &Store	&State &Department &Category &Total	&Total	&All series \\
    \midrule																	
    \texttt{Bottom-up}																	\\
    \hspace{0.1cm} 	SL	&SL	&None	&1.07 (0.00)	&420 (5.51)	&772 (12.94)	&70 (0.43)	&131 (0.84)	&312 (2.27)	&4.41 (0.01)	&2.2 (0.00)	&181 (1.45)	&343 (2.57)	&812 (7.46)	&1628 (24.81)	&2.2 (0.01)	\\
    \hspace{0.1cm} 	SL	&HL	&None	&1.07 (0.00)	&410 (5.82)	&743 (12.96)	&68 (0.37)	&128 (0.75)	&309 (1.79)	&4.39 (0.01)	&2.19 (0.00)	&177 (1.29)	&332 (2.25)	&808 (5.34)	&1653 (40.22)	&2.2 (0.01)	\\
    \hspace{0.1cm} 	HL	&HL	&None	&1.06 (0.00)	&342 (7.54)	&598 (18.24)	&63 (0.58)	&117 (1.36)	&293 (4.26)	&4.33 (0.01)	&2.18 (0.00)	&154 (2.01)	&283 (4.61)	&724 (14.31)	&1427 (81.18)	&2.1 (0.01)	\\
    \hspace{0.1cm} 	HL	&SL	&None	&1.06 (0.00)	&340 (6.2)	&597 (18.44)	&63 (0.56)	&117 (1.19)	&294 (3.06)	&4.33 (0.01)	&2.18 (0.00)	&154 (1.91)	&284 (4.49)	&725 (13.35)	&1417 (57.53)	&2.1 (0.01)	\\
    \hspace{0.1cm} 	TL	&HL	&None	&1.05 (0.00)	&350 (4.5)	&634 (10.36)	&65 (0.17)	&122 (0.49)	&299 (1.69)	&4.27 (0.01)	&2.15 (0.00)	&160 (0.69)	&305 (1.95)	&747 (9.07)	&1425 (50.6)	&2.1 (0.00)	\\
    \hspace{0.1cm} 	TL	&SL	&None	&1.06 (0.00)	&359 (4.18)	&647 (10.08)	&66 (0.25)	&124 (0.52)	&305 (1.54)	&4.3 (0.01)	&2.17 (0.00)	&164 (1.27)	&311 (2.84)	&778 (4.19)	&1457 (28.24)	&2.1 (0.00)	\\
    \hspace{0.1cm} 	TL	&TL	&None	&1.09 (0.00)	&866 (4.9)	&1843 (11.88)	&102 (0.44)	&210 (1.1)	&515 (4.95)	&5.03 (0.01)	&2.35 (0.00)	&306 (1.48)	&645 (3.75)	&1601 (17.55)	&4647 (59.66)	&2.8 (0.01)	\\
    \midrule																	
    \texttt{Sep. agg.}																	\\
    \hspace{0.1cm} 	SL	&SL	&Base	&1.07 (0.00)	&467 (13.86)	&819 (36.81)	&71 (0.62)	&144 (2.96)	&336 (9.36)	&4.23 (0.01)	&2.13 (0.00)	&187 (4.24)	&393 (11.11)	&971 (50.11)	&2523 (105.3)	&2.2 (0.00)	\\
    \hspace{0.1cm} 	SL	&SL	&OLS	&1.03 (0.00)	&455 (8.08)	&895 (10.88)	&70 (0.73)	&129 (0.87)	&311 (1.76)	&4.24 (0.01)	&2.13 (0.00)	&181 (1.81)	&345 (2.84)	&910 (9.07)	&2430 (45.28)	&2.2 (0.01)	\\
    \hspace{0.1cm} 	SL	&SL	&WLS-struct	&1.03 (0.00)	&417 (2.56)	&876 (8.99)	&64 (0.28)	&124 (0.66)	&298 (1.64)	&4.22 (0.00)	&2.12 (0.00)	&168 (1)	&334 (2.27)	&842 (6.11)	&2315 (22.33)	&2.1 (0.00)	\\
    \hspace{0.1cm} 	SL	&SL	&WLS-var	&1.04 (0.00)	&383 (3.33)	&771 (10.62)	&63 (0.25)	&120 (0.66)	&286 (1.48)	&4.24 (0.00)	&2.13 (0.00)	&162 (1)	&318 (2.75)	&770 (6.87)	&1887 (32.1)	&2.1 (0.00)	\\
    \hspace{0.1cm} 	SL	&SL	&MinT-shrink	&1.04 (0.00)	&393 (10.24)	&811 (27.92)	&62 (0.56)	&118 (1.64)	&283 (4.33)	&4.23 (0.01)	&2.13 (0.00)	&162 (2.44)	&316 (7.33)	&796 (18.35)	&2116 (81.24)	&2.1 (0.01)	\\
    \hspace{0.1cm} 	SL	&SL	&ERM	&1.26 (0.00)	&490 (28.29)	&935 (91.26)	&76 (1.99)	&139 (5.8)	&335 (17.11)	&5.26 (0.03)	&2.67 (0.01)	&201 (7.9)	&383 (22.22)	&966 (63.05)	&2559 (296.35)	&2.6 (0.04)	\\
    \midrule																	
    \texttt{Global}																	\\
    \hspace{0.1cm} 	SL	&SL	&Base	&1.12 (0.01)	&436 (22.31)	&900 (156.94)	&70 (1.01)	&133 (3.84)	&326 (15.88)	&4.37 (0.01)	&2.18 (0.01)	&188 (8.04)	&373 (13.48)	&1314 (611.36)	&2569 (320.87)	&2.3 (0.04)	\\
    \hspace{0.1cm} 	SL	&SL	&OLS	&1.05 (0.01)	&456 (32.4)	&906 (97.58)	&73 (11.48)	&139 (28.62)	&352 (93.72)	&4.38 (0.03)	&2.19 (0.04)	&197 (41.58)	&388 (103.17)	&1033 (336.46)	&2441 (389.89)	&2.3 (0.13)	\\
    \hspace{0.1cm} 	SL	&SL	&WLS-struct	&1.05 (0.00)	&456 (17.42)	&970 (53.68)	&68 (0.83)	&133 (2.1)	&323 (7.31)	&4.4 (0.01)	&2.18 (0.01)	&181 (3.3)	&368 (10.47)	&937 (40.46)	&2552 (258.94)	&2.2 (0.02)	\\
    \hspace{0.1cm} 	SL	&SL	&WLS-var	&1.06 (0.00)	&522 (18.96)	&1082 (59.2)	&76 (0.8)	&148 (3.13)	&353 (13.72)	&4.43 (0.02)	&2.19 (0.01)	&206 (4.75)	&415 (14.39)	&1017 (62.93)	&2744 (260.65)	&2.3 (0.02)	\\
    \hspace{0.1cm} 	SL	&SL	&MinT-shrink	&1.06 (0.01)	&474 (24.59)	&960 (92.69)	&74 (1.56)	&141 (3.13)	&358 (11.28)	&4.49 (0.06)	&2.22 (0.02)	&190 (7)	&367 (22.7)	&935 (87.38)	&2605 (424.61)	&2.3 (0.03)	\\
    \hspace{0.1cm} 	SL	&SL	&ERM	&1.25 (0.06)	&625 (60.87)	&1252 (148.15)	&90 (9.25)	&175 (24.13)	&426 (85.57)	&5.26 (0.29)	&2.65 (0.14)	&252 (34.4)	&512 (89.23)	&1285 (310.6)	&3190 (636.58)	&2.8 (0.19)	\\   
    
    \bottomrule    
  \end{tabular}}}
\end{sidewaystable*}

\end{document}